\documentclass[runningheads]{llncs}

\usepackage{eccv}

\usepackage{eccvabbrv}

\usepackage{graphicx}
\usepackage{booktabs}
\usepackage{amsmath}
\usepackage{amssymb}
\usepackage{booktabs}
\usepackage{multirow}
\usepackage{multicol}
\usepackage{makecell}
\usepackage{bm}
\usepackage{ltablex}

\usepackage[accsupp]{axessibility}  

\usepackage{hyperref}

\usepackage{orcidlink}

\def\mL{\mathcal{L}}

\def\mP{\mathcal{P}}

\def\1n{\mathbf{1}_n}
\def\0{\mathbf{0}}
\def\1{\mathbf{1}}

\definecolor{pink}{rgb}{0.9,0.5,0.5}
\definecolor{purple}{rgb}{0.5, 0.4, 0.8}   
\definecolor{gray}{rgb}{0.3, 0.3, 0.3}
\definecolor{mygreen}{rgb}{0.2, 0.6, 0.2}

\definecolor{greena}{rgb}{0.4, 0.5, 0.1}

\definecolor{bluea}{rgb}{0, 0.4, 0.6}

\definecolor{reda}{rgb}{0.6, 0.2, 0.1}

\newcommand{\cm}[1]{}

\newcommand{\myheading}[1]{\vspace{1ex}\noindent \textbf{#1}}

\newif\ifshowsolution
\showsolutiontrue

\ifshowsolution

\else

\fi

\newcommand{\Fref}[1]{Fig.~\ref{#1}}
\newcommand{\Tref}[1]{Table~\ref{#1}}

\begin{document}

\title{Look Hear: Gaze Prediction for \\ Speech-directed Human Attention} 



\titlerunning{Look Hear: Gaze Prediction for Speech-directed Human Attention}

\author{Sounak Mondal\inst{1} \and
Seoyoung Ahn\inst{2} \and
Zhibo Yang\inst{3} \and 
Niranjan Balasubramanian\inst{1} \and
Dimitris Samaras\inst{1} \and
Gregory Zelinsky\inst{1} \and
Minh Hoai\inst{4}}

\authorrunning{S. Mondal et al.}

\institute{Stony Brook University, NY, USA \and UC Berkeley, CA, USA \and Waymo LLC \and The University of Adelaide, Adelaide, Australia}

\maketitle

\begin{abstract}
For computer systems to effectively interact with humans using spoken language, they need to understand how the words being generated affect the users' moment-by-moment attention. Our study focuses on the incremental prediction of attention as a person is seeing an image and hearing a referring expression defining the object in the scene that should be fixated by gaze. To predict the gaze scanpaths in this \textit{incremental object referral} task, we developed the \textit{Attention in Referral Transformer} model or \textit{ART}, which predicts the human fixations spurred by each word in a referring expression. 
ART uses a multimodal transformer encoder to jointly learn gaze behavior and its underlying grounding tasks, and an autoregressive transformer decoder to predict, for each word, a variable number of fixations based on fixation history.  
To train ART, we created \textit{RefCOCO-Gaze}, a large-scale dataset of 19,738 human gaze scanpaths, corresponding to 2,094 unique image-expression pairs, from 220 participants performing our referral task. 
In our quantitative and qualitative analyses, ART not only outperforms existing methods in scanpath prediction, but also appears to capture several human attention patterns, such as waiting, scanning, and verification. Code and dataset are available at:  \url{https://sites.google.com/view/refcoco-gaze}.
\end{abstract}

\section{Introduction}
\label{sec:intro}

Humans are unique in that we use language to direct each others' attention in visual tasks. For example, a customer telling a baker ``I'd like the smallest pastry on the left'' communicates the desired object that needs to be selected. 
Understanding the human capacity to use these \textit{referring expressions} to incrementally direct attention is an important problem in cognitive science and has been studied for over half a century, with the more recent studies adopting eye-tracking methods~\cite{cooperControlEye1974, tanenhausIntegrationVisual1995, altmannLanguageCan2011, knoeferleVisuallySituated2016}. Most relevant is work showing the very tight link between a word in a referring expression and the very next eye movements of the person hearing it
~\cite{tanenhaus1996using, kamide2003time}, suggesting that humans 
\textit{incrementally} integrate visual information and word-by-word linguistic guidance in our attention control. However, these studies are limited in that they used small numbers of simple objects (often line drawings) in arrays (not scenes) and this constrained the linguistic complexity of the referring expression. How spoken language guides another person's attention in more naturalistic and ecologically valid contexts is still an open question in cognitive science. 

As our interactions with computers, vehicles, and AR/VR devices deepen, human-computer interaction (HCI) systems also need to give spoken guidance to users that is similarly effective in directing their attention. But, to attain this degree of synchrony with users, HCI systems must be able to integrate vision and language inputs to predict human gaze.  
Applications with this predictive ability will be highly time-sensitive and crucial for activities such as voice-assisted VR driving, offering a streamlined and immersive user experience where a person's gaze can be directed by generated spoken language as if the interaction were with another person.
 Being able to incrementally predict how a user integrates their visual input with a spoken instruction to direct their attention is a general advance in speech-assisted HCI, benefiting a broad range of applications, including 
efficient foveated rendering~\cite{pai2016gazesim}, VR sickness reduction~\cite{adhanom2020effect}, AR/VR eye–hand coordination analysis during human-object interaction~\cite{lavoie2024comparing}, VR skill training/assessment (\eg, driving~\cite{lang2018synthesizing}, surgery~\cite{bapna2023eye}), and user engagement analysis~\cite{khokhar2019eye}. Using predicted gaze for each word as guidance will enable speech-assisted HCI systems to incrementally generate efficient and clear instructions, correctly guiding user attention. Measuring gaze using eye-trackers instead of predicting it is more accurate, but also costlier and has limited applicability to aforementioned scenarios (such as time-crucial HCI) which require gaze prediction, and to situations where eye-trackers are unavailable or prohibited. 

In this context, we study the \textit{incremental object referral task}, for which we \textit{incrementally} predict eye movements of humans searching for a target object in an image as they are hearing a \textit{referring expression} describing that target. This task has not been studied previously in the context of human gaze prediction. The standard object referral task~\cite{mao2016generation, yu2016modeling} requires localizing the target object given an image and a referring expression. Our task is different in that we aim to incrementally predict the attention of a human as they are hearing the expression. Our task is also related to the categorical search task~\cite{zelinsky2020changing}, where humans direct their attention to a category of target object in an image. However, our incremental object referral task differs from categorical search in two key aspects. First, in incremental object referral, the target is designated by a complex referring expression (\eg, ``red baseball glove on the desk'') since the image may contain other objects belonging to the same category as the target. Referring expressions often refer to a target by its attributes (``red'') or its spatial relationships to other objects (``on the desk''), making it even more challenging to precisely localize the target without the broader descriptive context. In categorical search, the target is designated by only its category name (\eg, ``baseball glove''), which excludes the spatial terms and attributes commonly used by humans to describe an object. This task is also less realistic in that images depicting multiple instances of the target category are excluded (as spatial terms and attributes are not used). 
Second, because categorical search studies of attention use only an one or two word category name to designate a target (\eg, ``baseball glove''), most of the human search fixations occur only \textit{after} the complete referring ``expression'' is provided. This makes it an impoverished example of the longer and more natural referring expressions that we hope to study, ones requiring an incremental allocation of attention. 
Our more ecologically valid incremental object referral task therefore contributes to this cognitive science question by enabling exploration of natural referential expressions in real-world image contexts and generating testable hypotheses about how humans integrate language and vision.

Given its differences from related tasks that were studied previously, the incremental object referral task presents several technical challenges that necessitate more than mere updates to existing models. For instance, a standard object referral model cannot be re-purposed for incremental prediction. Although it is possible to sequentially input incomplete referring expressions, one word at a time, and use the predicted positions as proxies for fixation locations, this approach proves ineffective as existing object referral models train on complete referring expressions in contrast to how humans integrate visual and linguistic information in a word-by-word basis~\cite{tanenhaus1996using, kamide2003time}. Alternatively, one can adapt an existing scanpath prediction model for our incremental object referral task, but this approach is also unsuitable because existing scanpath prediction models do not learn the object grounding processes (for \textit{both} partial and complete referring expressions) hypothesized to underlie the gaze behavior observed in our task.


To address the aforementioned challenges of the incremental object referral task, we introduce the \textit{Attention in Referral Transformer (ART)} model. ART is tailored to the multimodal demands of our task as it uses a multimodal transformer encoder that jointly learns gaze prediction and object grounding objectives. Furthermore, we integrate an autoregressive transformer decoder that leverages fixation history to better predict the subsequent fixations corresponding to each sequentially presented word from the referring expression. This innovative decoder component flexibly adjusts both the count and the parameters of predicted fixations in alignment with the evolving input, thus mirroring the dynamic nature of human attention.


\begin{figure}[tb]
  \centering
  \def\subFigSz{0.32\linewidth}
\centering
\includegraphics[width=\subFigSz]{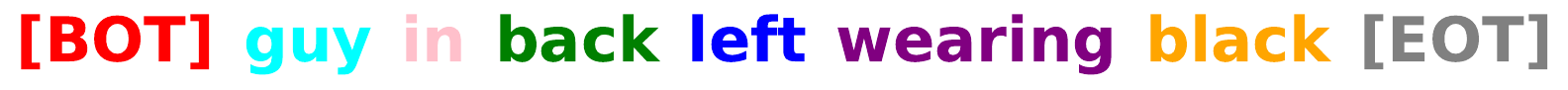} 
\includegraphics[width=\subFigSz]{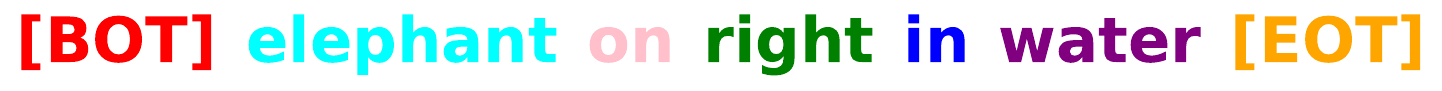} \includegraphics[width=\subFigSz]{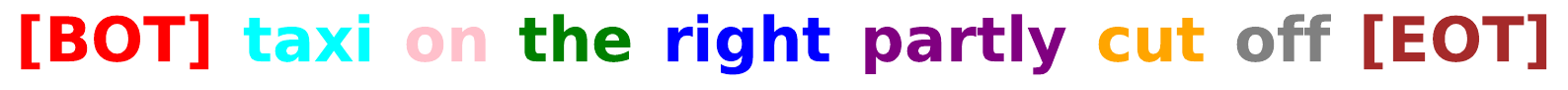}\\
\includegraphics[width=\subFigSz]{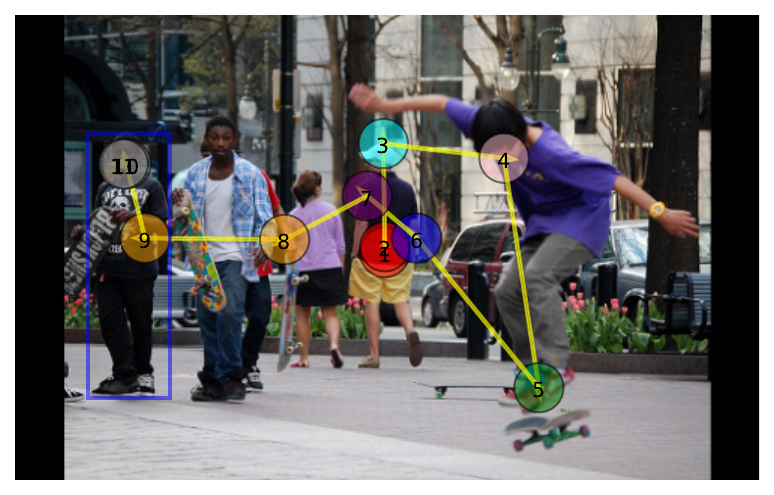} 
\includegraphics[width=\subFigSz]{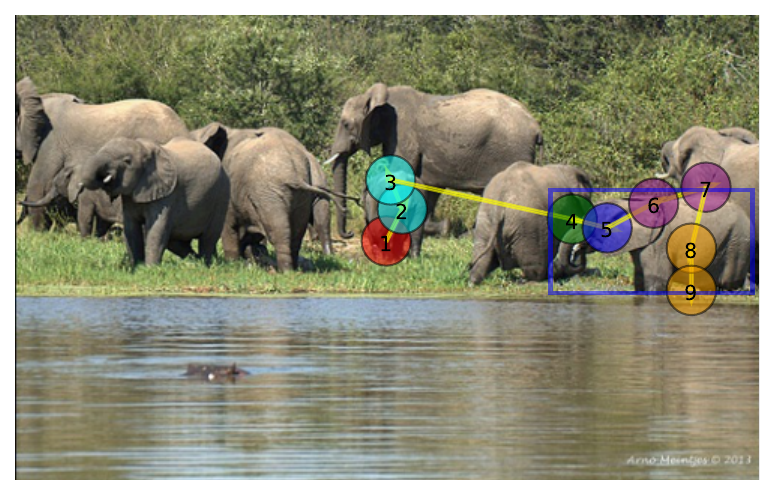} \includegraphics[width=\subFigSz]{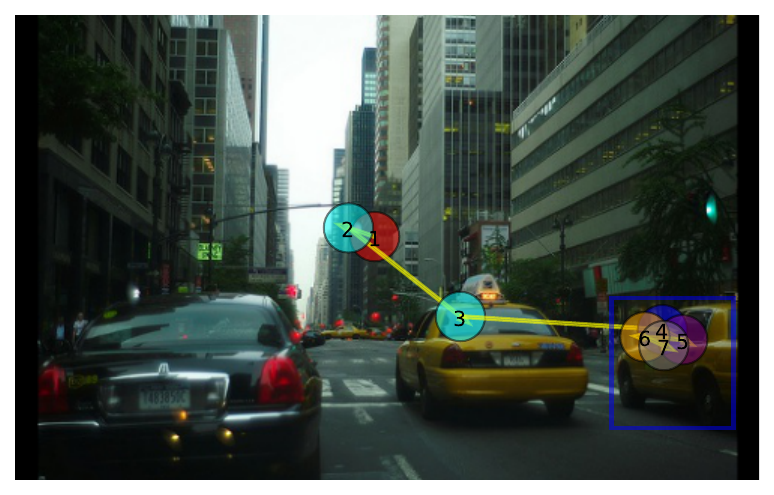} 
\caption{\textbf{RefCOCO-Gaze Dataset.} Sample image-expression pairs and corresponding scanpaths under our \textit{incremental object referral} task. Fixations (denoted by circles numbered with fixation order) are color-coded to the corresponding word in the referring expression (above each image). Fixations color-coded to \texttt{[BOT]} occurred before the expression started, and fixations color-coded to \texttt{[EOT]} occurred after the expression ended. Blue bounding boxes indicating referred objects were not visible during trials. 
  }
  \label{fig:teaser}
\end{figure}

To train ART, we collected \textit{RefCOCO-Gaze} (Fig.~\ref{fig:teaser}), a large-scale dataset of gaze behavior from 220 people performing the incremental object referral task on 2,094 images and associated referring expressions from RefCOCO~\cite{yu2016modeling} dataset. Compared to baselines~\cite{chen2021predicting, wang2022ofa, mondal2023gazeformer}, only ART was able to accurately predict the dynamic changes in gaze of humans incrementally hearing and shifting their attention in response to the words in the referring expression, even with incomplete target descriptions. Ablation studies showed that pre-training and training on auxiliary grounding tasks, such as object localization and target category prediction, improves ART's gaze prediction performance. In qualitative analyses, ART was shown to capture several fixation patterns of people performing the incremental object referral task, such as waiting, scanning, and verification, suggesting that it learned to strategically disambiguate vision-language ambiguities. 

In summary, our contributions are: \textbf{(1)} Introducing the \textit{incremental object referral} task for gaze prediction that will lead to more user-responsive HCI systems. \textbf{(2)} Creating \textit{RefCOCO-Gaze}, a large-scale dataset of gaze behavior during the incremental object referral task. \textbf{(3)} Developing \textit{ART}, the first gaze prediction model of incremental object referral that offers computational solutions to the incremental and multimodal aspects of our task. \textbf{(4)} Bringing RefCOCO-Gaze and ART 
into the toolboxes of researchers studying incremental object referral, thereby enabling them to understand how humans dynamically merge their visual and linguistic information in the real world to control their attention.


\section{Related Work}
\label{sec:related_work}

Interest in gaze prediction as a computer vision problem has been growing~\cite{yang2020predicting, chen2021predicting, mondal2023gazeformer,Yang_2024_CVPR}, given that the anticipation of user attention would enable more natural augmented/virtual reality systems~\cite{pai2016gazesim, chung2022static, bennett2021assessing,min2021integrating}. Most existing human attention prediction models predict free-viewing behavior
~\cite{IttiPAMI98, masciocchi2009everyone, berg2009free}, but fail to generalize to goal-directed behaviors, such as visual search~\cite{henderson2007visual, koehler2014saliency}. More related is the work predicting eye fixations during the search for a target object~\cite{zelinsky2019benchmarking,zelinsky2021predicting}. Another study~\cite{yang2020predicting} 
predicted 
fixation scanpaths during search using COCO-Search18~\cite{chen2021coco}, a dataset of search fixations. Using a dataset~\cite{chen2020air} of fixation scanpaths from a Visual Question Answering (VQA) task, Chen \etal~\cite{chen2021predicting} proposed a model that predicted both VQA and search behavior, and both Chen \etal~\cite{chen2022characterizing}, Yang \etal~\cite{yang2022target} proposed models to predict both target-present and target-absent search. Mondal \etal~\cite{mondal2023gazeformer} proposed a multimodal transformer model called Gazeformer, which achieves state-of-the-art search prediction performance while generalizing well to unknown targets. Gazeformer~\cite{mondal2023gazeformer} and Chen \etal~\cite{chen2021predicting} can be adapted for our multimodal task, and serve as baselines in Sec.~\ref{section:experiments}.

Despite this increasing interest in gaze prediction as a computer vision problem, no existing model effectively addresses the incremental object referral task. 
Large vision-language  foundation models~\cite{radford2021learning, jia2021scaling, yu2022coca, wang2022ofa, yuan2021florence, alayrac2022flamingo} yield unprecedented performance in visual, lingual and cross-modal tasks and effectively generalize to new concepts and tasks. Moreover, several tasks simply require multimodal modeling, such as VQA~\cite{vqa2015antol, mensink2023encyclopedic}, image captioning~\cite{fang2015captions, kuo2022beyond}, and object referral~\cite{yu2016modeling, mao2016generation, yan2023universal}. Relatedly, \textit{object referral} (also known as visual/object grounding of referring expressions) localizes or \textit{grounds} a single unambiguous object in an image that is referred to in a natural language \textit{referring expression}. Thus, the input is an image-text pair and the output is the referred object's bounding box parameters. Recent object referral models have adopted two-stage~\cite{hong2019learning,hu2017modeling, liu2019learning} and one-stage~\cite{chen2018real, deng2021transvg, liao2020real} architectures, 
and to this end, several high-quality object referral benchmarks have been curated, such as ReferItGame~\cite{kazemzadeh2014referitgame}, RefCOCO~\cite{yu2016modeling}, RefCOCO+~\cite{yu2016modeling}, and RefCOCOg~\cite{mao2016generation}. 
 Researchers have also studied human attention as people view an image and concurrently describe it~\cite{vaidyanathan2018snag, vaidyanathan2020computational}. One study~\cite{pont2020connecting} collected a dataset of spoken image descriptions where each word was visually grounded by a mouse trace. He \etal~\cite{he2019human} collected a dataset containing fixations (recorded by an eye-tracker) synchronized with concurrently spoken image descriptions. However, these studies specifically focused on \textit{spoken description of the entire image} and not object referral. Vasudevan \etal~\cite{vasudevan2018object1} explored object referral for previously spoken referring expressions, and did not predict human attention. Another study~\cite{vasudevan2018object} on spoken object referral in videos used human gaze and spoken referring expression as inputs. Zhang \etal~\cite{zhang2022gaze} collected a dataset of static gaze estimation heatmaps for non-incremental referral. To our knowledge, we are the first to computationally model human gaze and explore its interactions with vision and language in a realistic incremental referral task. 

\begin{figure}[tb]
  \centering
  \includegraphics[width=0.9\textwidth]{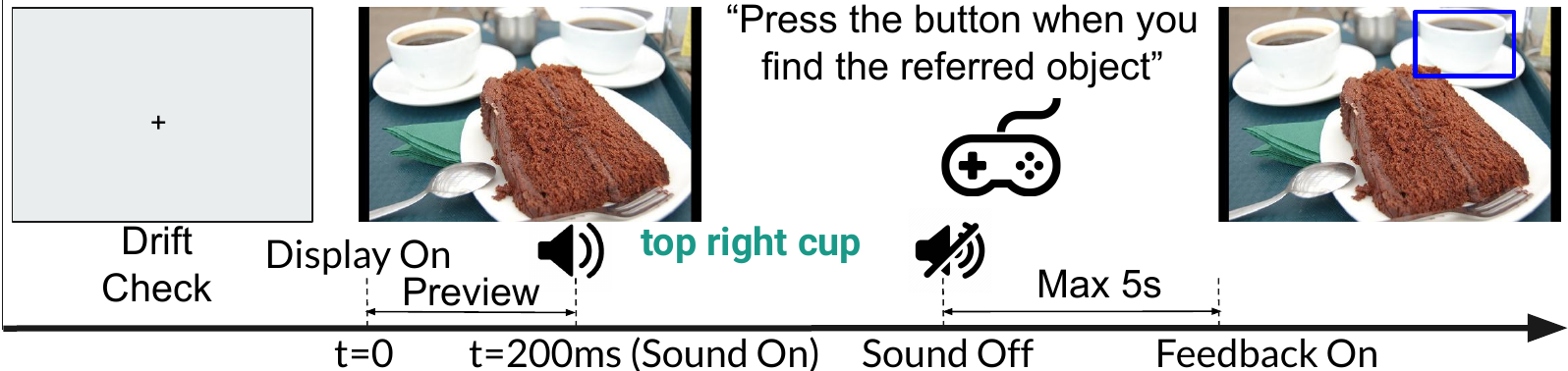}
  \caption{Behavioral data collection in our incremental object referral paradigm.}
  \label{fig:task}
\end{figure}
\section{RefCOCO-Gaze Dataset}
\label{sec:dataset}

RefCOCO-Gaze is the largest dataset for studying human gaze behavior during an incremental object referral task. It consists of 19,738 scanpaths that were recorded while 220 participants with normal or corrected-to-normal vision viewed 2,094 COCO~\cite{linMicrosoftCoco2014} images and listened to the associated referring expressions from the RefCOCO dataset~\cite{yu2016modeling}. RefCOCO was collected using the ReferItGame~\cite{kazemzadeh2014referitgame} where players must construct efficient referring expressions for another player to locate the correct object. RefCOCO mirrors real-life speech which is known to contain elliptical and unstructured expressions~\cite{townendwalker2006, thanh2015differences}. 
The gaze data, recorded by an EyeLink 1000 eyetracker,  includes information about the location and duration of each fixation, the bounding box of the search target, audio recordings of the referring expressions, the timing of the target word, and the synchronization between the spoken words and the sequence of fixations (tells us which word triggered which fixations). RefCOCO-Gaze covers a diverse range of linguistic and visual complexity, making it an ideal dataset for researchers studying human integration of vision and language, and HCI researchers alike. 


  

\Fref{fig:task} depicts the incremental object referral paradigm used for human gaze collection. 
We have selected 2,094 image-expression pairs from the larger RefCOCO dataset based on the target object size, image ratio, and sentence complexity. 
Each participant performed ${\sim}100$ trials
, yielding 10-16 scanpaths per image-expression pair. Participants were instructed to move their gaze as quickly as possible to the target object that is being referred to in a language expression played through a speaker. Each trial began with drift correction (for accurate eye tracking) and presentation of an image. The image was displayed for 200ms before the audio onset (too short a time for an eye movement). 
The image remained visible until the participant pressed a button to indicate that they found the target or until five seconds elapsed following completion of the auditorily presented referring expression. At the end of each trial, the correct bounding box location of the target object was provided as feedback, followed by a survey asking whether the participant indeed found and recognized the target. A target was present in each image. We used a forced aligner~\cite{mcauliffe2017montreal}, a tool for aligning speech with text, to synchronize gaze movements with individual words of a referring expression. This study had IRB approval.

We divided RefCOCO-Gaze into disjoint training and evaluation sets that preserve the approximate proportion of training to evaluation data in RefCOCO. The training set consisted of 1799 image-expression pairs (selected \textit{only} from the original RefCOCO train split), corresponding to 16,982 scanpaths. Scanpaths from image-expression pairs from validation and test splits of RefCOCO were randomly shuffled and split (1:2 ratio) to create disjoint validation (92 image-expression pairs, 869 scanpaths) and test (203 image-expression pairs, 1887 scanpaths) sets -- both having a balanced distribution of target categories. Dataset details (\eg, stimuli selection, gaze recording, pre-processing, comparison with related datasets, etc.) and extensive dataset analyses are in the supplement.

\begin{figure}[tb]
\centering
\includegraphics[width=\linewidth]{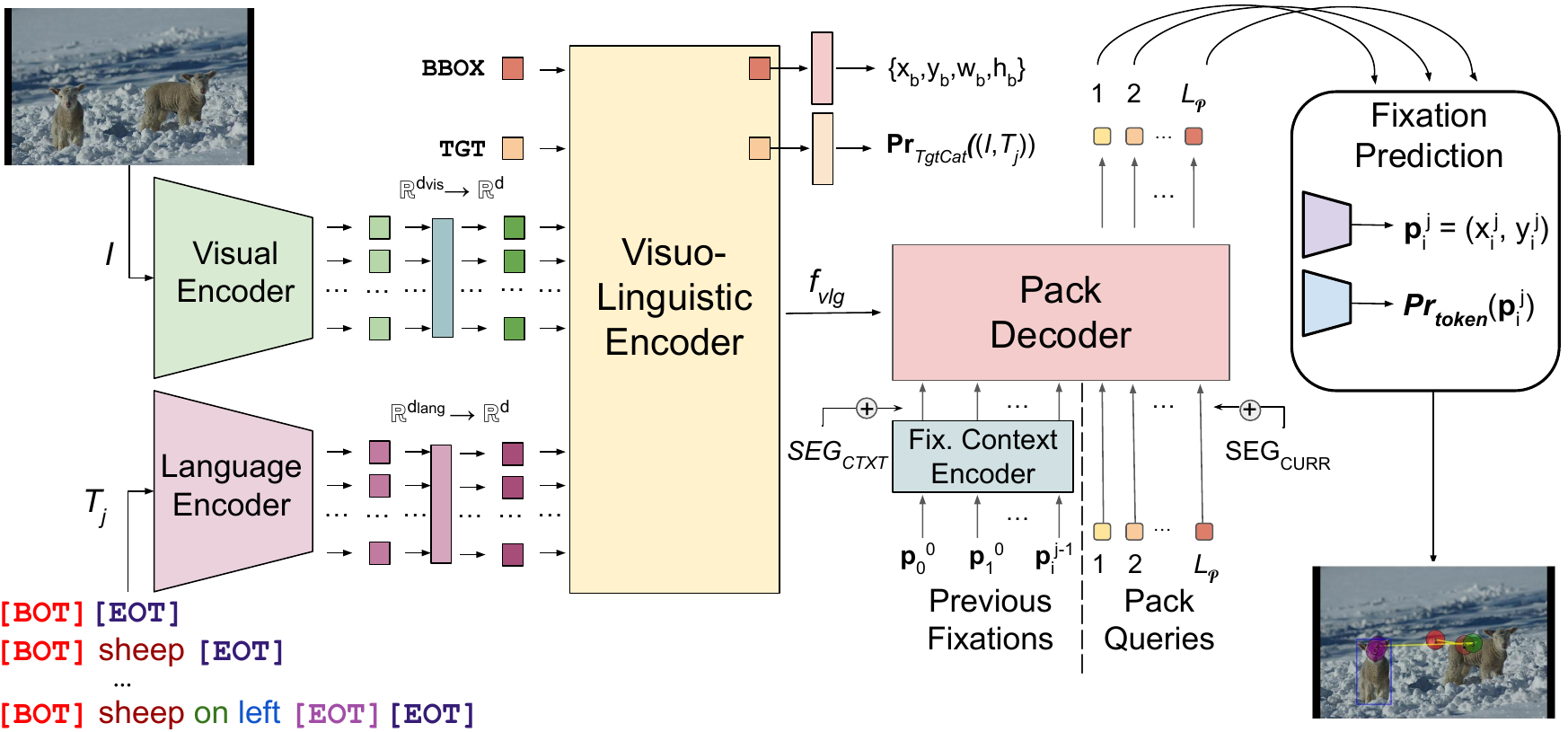}
%

\caption{{\bf Attention in Referral Transformer (ART) Architecture.} On each pass after comprehending a new word, the model takes an image $I$ and tokens $T_j$ of prefix $R_j$ of the referring expression as input and generates a possibly empty sub-sequence of fixations based on previous fixation history encoded by a fixation context encoder. }
\label{fig:arch}
\end{figure}

\section{Gaze Prediction for Incremental Object Referral}
\label{sec:method}
Our goal is to predict fixations as a person progressively receives information about the referred object through \textit{each word} of the referring expression that they are hearing. We design a novel multimodal transformer architecture called \textit{\textbf{A}ttention in \textbf{R}eferral \textbf{T}ransformer} or \textbf{ART}  for this task. ART solves multiple problems that arise when adapting previous gaze prediction models~\cite{mondal2023gazeformer, chen2021predicting} for the incremental object referral task, owing to its several novel features that these models~\cite{mondal2023gazeformer, chen2021predicting} lack: 
\textbf{(1)} ART integrates an object referral network into our gaze prediction framework and trains it on partial expressions, whereas previous baselines either extract task guidance from a frozen object referral model~\cite{kamath2021mdetr} trained on complete expressions~\cite{chen2021predicting}, or lack object grounding capabilities~\cite{mondal2023gazeformer}. \textbf{(2)} ART integrates a novel fixation prediction framework (absent in previous methods) that accommodates the autoregressive prediction of any number of fixations triggered by a word -- zero or one or several, based on previous fixations encoded by a novel fixation context encoder (in contrast to non-autoregressive Gazeformer~\cite{mondal2023gazeformer} which predicts fixations with the knowledge of the first fixation alone). \textbf{(3)} ART is an object referral network with a scanpath prediction decoder module, which allows us to \textit{pre-train} the object referral network on a large-scale object referral dataset~\cite{yu2016modeling} using object grounding objectives, thus generalizing better despite training on a much smaller gaze dataset. ART also jointly trains on the object grounding objectives along with the primary fixation prediction objective. On the other hand, previous baselines either rely on frozen object referral models~\cite{chen2021predicting} or lack an object grounding subnetwork altogether~\cite{mondal2023gazeformer} and are therefore limited to only training on the fixation prediction objective alone. 

\subsection{Architecture}

The overall architecture of ART is shown in \Fref{fig:arch}. Since the understanding of the referred object changes with each incoming word, we design ART to output a possibly empty sequence of fixations (which we dub a ``pack'' of fixations) triggered by a new word $w_j$ from the referring expression. We follow single-stream vision-language model architectures~\cite{deng2021transvg, li2019visualbert, li2022toist,li2023understanding} to design a multimodal transformer encoder module that encodes visuo-linguistic context for object referral. 
A transformer decoder module generates embeddings for incremental scanpath prediction conditioned by visuo-linguistic context from the encoder and fixation history encoded by a fixation context encoder module.

\myheading{Visual and Language Encoders.} We use separate encoders for the vision and language modalities. As in DETR~\cite{carion2020end}, the \textbf{visual encoder} consists of a ResNet-50~\cite{he2016deep} backbone followed by a standard transformer encoder module~\cite{vaswaniAttention}. Given an image $I\in \mathbb{R}^{3{\times} H{\times} W}$, the visual encoder generates patch embeddings $g_{vis}\in \mathbb{R}^{d_{vis}{\times}hw}$. A \textbf{language encoder} (RoBERTa~\cite{liu2019roberta}) encodes a \textit{prefix} $R_j=\{w_1,\ldots, w_j\}$ upon utterance of the $j^{th}$ word $w_j$ in a referring expression. $R_j$ is tokenized~\cite{sennrich2016neural} to obtain $T_j=\{\texttt{[BOT]}, t_1, ..., \texttt{[EOT]}\}$, which is then processed by the language encoder to yield language embedding sequence $g_{lang} \in \mathbb{R}^{d_{lang}{\times} l_{lang}}$. Here, $l_{lang}$ is the maximum number of tokens that can be processed by the language model; $\texttt{[BOT]}$ and $\texttt{[EOT]}$ are the beginning-of-text and end-of-text tokens, respectively. To predict fixations triggered \textit{before} the first word has been spoken, the input tokens to the language encoder is the sequence $\{\texttt{[BOT]},\texttt{[EOT]}\}$, and to predict fixations \textit{after} the  expression has ended, we append an additional $\texttt{[EOT]}$ token to the tokenized input. Existing baselines ~\cite{mondal2023gazeformer, chen2021predicting} use ResNet-50 and RoBERTa, hence we choose them for fair comparison.

\myheading{Visuo-linguistic Transformer Encoder.} 
In contrast to Gazeformer~\cite{mondal2023gazeformer}, which utilized pooled text encodings from a frozen language encoder for task description and risked losing word-level details, our method integrates both image and linguistic tokens into a unified sequence of multimodal tokens, thus allowing fine-grained word-level interactions between image and linguistic tokens. We process this sequence through a visuo-linguistic encoder, which is designed as a standard transformer encoder~\cite{vaswaniAttention}. Given the set of patch embedding vectors \( g_{vis} \) and the sequence of language embedding vectors \( g_{lang} \) with different dimensionalities, as described above, we first project them to an embedding space of the same dimension $d$ using modality-specific projections to obtain $f_{vis}\in\mathbb{R}^{d{\times}hw}$ and $f_{lang}\in\mathbb{R}^{d{\times}l_{lang}}$. 
We also introduce two learnable $d$-dimensional embeddings $\texttt{BBOX}, \texttt{TGT}$ which correspond to the object localization and target category prediction tasks, respectively. The input to the  visuo-linguistic encoder is the concatenation of $\texttt{BBOX}$, $\texttt{TGT}$, $f_{vis}$, and $f_{lang}$. The corresponding output is tensor $f_{vlg}\in\mathbb{R}^{d{\times}(hw+l_{lang}+2)}$. 
We project $f_{vlg}[0]$ (corresponding to $\texttt{BBOX}$) to $\{x_b,y_b,w_b,h_b\}$ where $x_b,y_b$ are the coordinates of the upper-left corner, and $w_b$ and $h_b$ are the width and height of the bounding box. We also use a linear layer along with softmax to project $f_{vlg}[1]$ (corresponding to $\texttt{TGT}$) to the probability distribution $\mathbf{Pr}_{\textit{TgtCat}}$ over all possible target categories. 

\myheading{Per-Word Fixation Prediction Framework.} We predict fixations (or absence thereof) triggered by a new word of a referring expression containing $L$ words. Let the scanpath $\mathcal{S}$ for incremental object referral be a sequence of \textit{packs} of fixations. A pack $\mP_j = \{(x_{i}^{j}, y_{i}^{j})| i=0, 1, \ldots \}$ is an ordered sequence of 2D fixations, triggered by the change in knowledge about the referred object due to a new word $w_j$, for $j \in\{1,..,L\}$. While a pack is usually spurred by a word, a pack of fixations $\mathcal{P}_0$ can be triggered \textit{before} the first word is spoken, similar to free-viewing behavior. A pack of fixations $\mathcal{P}_{L+1}$ can also occur \textit{after} the referring expression has ended. A word may not inspire any fixations at all, yielding a \textit{null} pack $\mathcal{P}_\phi = \phi$. We also define a \textit{terminal} pack $\mathcal{P}_{TERM}$ which, like the null pack, does not contain any valid fixations, and denotes the end of scanpath (EOS). 

\myheading{Fixation context encoder.} We parameterize a fixation $\mathbf{p}^k_i$ using four parameters: $x$-location $x^k_i$, $y$-location $y^k_i$, the pack number $k$ (i.e., the index of the pack the fixation belongs to), and the within-pack index $i$ (which we call \textit{order}). We use this parameterization to capture the \textit{fixation context}, which refers to the information of previous fixations in the ongoing scanpath. We use a fixed 2D sinusoidal positional embedding~\cite{carion2020end}  to encode the spatial $x,y$ location to $XY_i^k\in\mathbb{R}^{2d}$ and two fixed 1D sinusoidal positional embeddings~\cite{vaswaniAttention} to encode the pack number $j$ and the order $i$ to $n_i^k, o_i^k\in\mathbb{R}^{d}$ respectively. $XY_i^k$, $n_i^k$, and $o_i^k$ are concatenated and projected to fixation encoding $\mathbf{c}_i^k\in \mathbb{R}^d$. Hence, for a new word $w_j$, we can construct an ordered sequence of 
$\mathbf{c}_i^k $ ($k < j, i=0,1,...$) and zero-pad to maximum length $L_{\mathcal{C}}$ to obtain the fixation context tensor $\mathcal{C}_{j}\in\mathbb{R}^{d{\times} L_{\mathcal{C}}}$.

\myheading{Pack decoder.} To obtain the current pack of fixations, we use a transformer decoder module~\cite{vaswaniAttention}. Let the input $\mathcal{Q}=\{q_k|k=1, \ldots, L_{\mathcal{P}}\}$ to the decoder be a sequence of pack queries $q_k$, where $L_{\mathcal{P}}$ is the maximum number of fixations in a pack and $q_k$'s are learnable vectors (similar to fixation queries \cite{mondal2023gazeformer}). To help the model differentiate the nature of context and pack embeddings, we further add two separate segment embeddings~\cite{devlin-etal-2019-bert}, namely $SEG_{ctxt}$ and $SEG_{curr}$ to previous fixation context tensor $\mathcal{C}_{j}$ and $\mathcal{Q}$, respectively. Next, we  use the concatenation of $\mathcal{C}_{j}$ and $\mathcal{Q}$ as input to the decoder. The decoder also receives dynamic visuo-linguistic context through cross-attention with $f_{vlg}$. The output from the decoder is tensor $f_{decoder}\in\mathbb{R}^{d{\times}(L_{\mathcal{C}}+L_{\mathcal{P}})}$. The last $L_{\mathcal{P}}$ $d$-dimensional slices of $f_{decoder}$ corresponding to the $L_{\mathcal{P}}$ pack queries are denoted as $f_{pack}\in\mathbb{R}^{d{\times} L_{\mathcal{P}}}$. 

\myheading{Fixation Prediction Module.} 
Fixation prediction for incremental object referral is challenging since a pack can have between zero and multiple fixations, and a scanpath can be terminated before the end of a referring expression. We account for these scenarios by making all packs be of length $L_{\mathcal{P}}$ with the following parameterization. First, any valid fixation in a pack is represented by a fixation token $\texttt{FIX}$. Second, null packs and packs having less than $L_{\mathcal{P}}$ valid fixations are padded with padding tokens $\texttt{PAD}$ to maximum length $L_{\mathcal{P}}$. 
 Third, we complete a terminal pack $\mathcal{P}_{TERM}$ with $L_{\mathcal{P}}$ termination tokens $\texttt{EOS}$. For each of the $L_{\mathcal{P}}$ slices of $f_{pack}$, we use a token prediction MLP 
and a softmax layer to predict if that slice corresponds to one of $\texttt{FIX}$, $\texttt{PAD}$, and $\texttt{EOS}$ tokens. We use regression heads and Gaussian distributions~\cite{mondal2023gazeformer} to model the fixation locations. We also augment ART with fixation duration modeling and detail it in the supplement. 


\subsection{Pre-training, Training, and Inference}

\myheading{Pre-training.} Since object grounding is at the core of our task, we \textit{pre-train} the visual, language and visuo-linguistic encoder modules on the two objectives that we hypothesize underlie the object grounding process: object localization and target category prediction, using RefCOCO~\cite{yu2016modeling} training data.  \textbf{Object localization} is the estimation of the referred object bounding box. 
We apply an $L_1$ regression loss $\mathcal{L}_{reg}$ and a generalized IoU (GIoU) loss  \cite{rezatofighi2019generalized} $\mathcal{L}_{giou}$, between predicted and  ground truth bounding box parameters. The \textbf{target category prediction} task discerns the object type from the expression (e.g., predict ``car'' in the expression ``left sedan next to the motorcycle''). We pre-train on this task using a cross-entropy loss $\mathcal{L}_{target}$. Total pre-training loss is
$\mathcal{L}_{pretrain} = \mathcal{L}_{reg}+ \mathcal{L}_{giou} + \mathcal{L}_{target}$. 

\myheading{Training \& Inference.} We train ART using the teacher-forcing algorithm~\cite{williams1989learning}, i.e., we provide the ground truth fixations to construct the fixation context and treat each pack in the training scanpaths as independent minibatch items. To train ART on the gaze prediction task, we apply $\mathcal{L}_1$ regression loss (following \cite{mondal2023gazeformer}) on the predicted $x$ and $y$ locations. Let the predicted pack of fixations $\mP_k = \{(x_{i}^{k}, y_{i}^{k})\}_{i=1}^{L_{\mathcal{P}}}$, and ground-truth pack of fixations $\hat{\mP}_k = \{(\hat{x}_{i}^{k}, \hat{y}_{i}^{k})\}_{i=1}^{l^{k}}$ where  $l^{k}$ is the length of the ground truth pack. Moreover, let $\hat{v}_{i,t}^{k}$ be a binary scalar representing ground truth of the $i^{th}$ token in $\mathcal{P}_k$ belonging to the token class $t\in T \text{ where } T = \{\texttt{FIX}, \texttt{PAD}, \texttt{EOS}\}$. Also let $v_{i,t}^{k}$ be the probability of that token belonging to token class $t$ as estimated by our model. The multitask loss for a minibatch of size $M$ is
\begin{align}
   & \mathcal{L}_{gaze} = \frac{1}{M} \sum_{k=1}^{M} \left( \mathcal{L}_{xy}^{k} + \mathcal{L}_{token}^{k} \right).
\end{align}
Here $\mathcal{L}_{xy}^{k} = \frac{1}{l^{k}}  \sum_{i=1}^{l^{k}} \left( |x_i^{k}  - \hat{x}_i^{k}| + |y_i^{k}  - \hat{y}_i^{k}|\right)$, $\mathcal{L}_{token}^{k} = - \sum_{i=1}^{L_{\mathcal{P}}} \sum_{t\in T} \hat{v}_{i,t}^{k}\log(v_{i,t}^{k})$. In addition to the gaze-prediction loss $\mathcal{L}_{gaze}$, we also train on the object localization and target category prediction tasks, but only \textit{after} either of the following two events has occurred: (1) the last word of the referring expression has been uttered, (2) the ground truth scanpath has been terminated. Note that both events ensure sufficient information in the referring expression comprehended thus far for a human to localize the object. This multi-task grounding loss $\mathcal{L}_{ground}$ is $\mathcal{L}_{bbox} + \mathcal{L}_{target}$, where $\mathcal{L}_{bbox} =  \mathcal{L}_{reg}+ \mathcal{L}_{giou}$.
Hence, the total multi-task loss $\mathcal{L}$
that we use to train our ART model is $\mathcal{L} = \mathcal{L}_{gaze} + \mathcal{L}_{ground}$ when the scanpath has terminated or the referral audio has ended, and $\mL = \mL_{gaze}$ otherwise. 
During inference, ART \textit{autoregressively} generates packs of fixations conditioned on the previous fixations generated by the model and the scanpath is terminated upon encountering the first termination token $\texttt{EOS}$ in a predicted pack. The fixations within a pack are efficiently generated in parallel.

\section{Experiments}
\label{section:experiments}

Here, we experimentally evaluate scanpath prediction capability for incremental object referral. For the conventional scanpath prediction task, accurately predicting the entire sequence of fixations is the main objective. However, for our task, it is perhaps equally important, if not more, for the predicted scanpath to be correct at the word-level granularity, i.e., packs (including null packs) must be predicted accurately. 
Following previous work~\cite{yang2020predicting, yang2022target, chen2021predicting, mondal2023gazeformer}, we sample 10 scanpaths per image-expression pair for all models. 
More details of ART, such as its design and implementation details, are in the supplement.

\subsection{Performance Metrics}

We use a broad set of metrics to evaluate dynamic word-based scanpath prediction for incremental object referral. \textit{Sequence Score} metric \cite{yang2020predicting} converts predicted and ground truth scanpaths into strings of fixation cluster IDs and compares them using a string matching algorithm \cite{needleman1970general}. \textit{Fixation Edit Distance}~\cite{mondal2023gazeformer} measures scanpath dissimilarity using the Levenshtein algorithm~\cite{Levenshtein1965BinaryCC} after converting scanpaths to strings like Sequence Score does.
We measure Sequence Score and Fixation Edit Distance in two granularities: (1) over the entire scanpath ($SS$ and $FED$); and (2) over a pack ($SS_{pack}$ and $FED_{pack}$), where $SS_{pack}$ and $FED_{pack}$ are the averages of sequence scores and fixation edit distances, respectively, between the ground-truth and predicted packs. 
We also introduce $CC_{pack}$ and $NSS_{pack}$, the word-based versions of the \textit{Correlation Coefficient (CC)~\cite{jost2005assessing}} and \textit{Normalized Scanpath Saliency (NSS)~\cite{peters2005components}} metrics. CC is the correlation between the normalized model saliency map and a Gaussian-convolved human fixation map. NSS averages the values of a model's fixation map at the locations fixated by humans~\cite{bylinskii2019different}, and is a discrete version of CC. $CC_{pack}$ and $NSS_{pack}$ are the averages of $CC$ scores and $NSS$ scores, respectively, over all possible packs. Higher $SS$, $SS_{pack}$, $NSS_{pack}$, and $CC_{pack}$ values signify more similarity between model-generated and human scanpaths, whereas lower $FED$ and $FED_{pack}$ scores indicate higher similarity. 
 More details are in the supplement.




\subsection{Baselines}

We compare ART with: (1) \textbf{Random Scanpath}: We uniformly sample a pack length value $l_p$ and then uniformly sample $l_p$ fixation locations from the image. 
(2) \textbf{OFA}: We use the state-of-the-art vision-language model OFA~\cite{wang2022ofa}, trained on several multimodal benchmarks. 
We uniformly sample pack length $l_p$ and then sample $l_p$ fixation locations within the OFA-predicted \textit{bounding box} for each referring expression prefix. (3) \textbf{Chen \etal}~\cite{chen2021predicting}:  This model learns goal-directed human gaze through a dynamically updated memory which is initialized by task guidance maps. To extend this model to our task, we create task guidance maps using bounding boxes from the SOTA referral model MDETR~\cite{kamath2021mdetr}, trained only on RefCOCO. 
(4) \textbf{Gazeformer-ref}: This baseline, based on Gazeformer~\cite{mondal2023gazeformer}, takes expression prefixes as target information and generates packs of fixations. 
(5) \textbf{Gazeformer-cat}: Since target category information might get lost in the pooled linguistic embedding used by Gazeformer-ref, we evaluate another variant of Gazeformer~\cite{mondal2023gazeformer} called \textit{Gazeformer-cat} which takes the  \textit{target category name} estimated for an expression prefix as input and treats the problem as categorical visual search. The target category estimation of a prefix is done by a pre-trained RoBERTa-based classifier. 
Find more details of the baselines in the supplement.

\setlength{\tabcolsep}{2pt}
\begin{table}[tb]
\centering
\caption{Performance of ART and baselines on RefCOCO-Gaze test set. }
\label{table:main_results}

\begin{tabular}{lcccccc}
\toprule 
& $SS\bm{\uparrow}$ & $SS_{pack}\bm{\uparrow}$& $FED\bm{\downarrow}$ & $FED_{pack}\bm{\downarrow}$&$CC_{pack}\bm{\uparrow}$& $NSS_{pack}\bm{\uparrow}$\\  \midrule 
Human & 0.400 & 0.317 & 6.573 & 1.278 & 0.283 & 3.112\\ \midrule
Random &  0.189 & 0.133 &  17.735 & 3.005 & 0.094 & 1.689\\
OFA~\cite{wang2022ofa} &  0.216 & 0.170 &  17.084 & 2.901 & 0.174 & 2.175 \\
Chen \etal~\cite{chen2021predicting} &  0.299 & 0.188 &  8.309 & 1.507 & 0.159 & 1.557\\
Gazeformer-ref~\cite{mondal2023gazeformer} &  0.269 & 0.194 &  6.788 & 1.286 & 0.208 & 3.006\\
Gazeformer-cat~\cite{mondal2023gazeformer} &  0.269 & 0.189 &  6.841 & 1.327 & 0.204 & 2.932\\ 
ART (Proposed) &  \textbf{0.359} & \textbf{0.292} &  \textbf{6.371} & \textbf{1.143} & \textbf{0.280} & \textbf{3.478}\\
\bottomrule
\end{tabular}

\end{table}

\def\subFigSz{0.23\linewidth}
\begin{figure}[tb]
\centering

\includegraphics{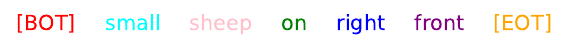} \\
\includegraphics[width=\subFigSz]{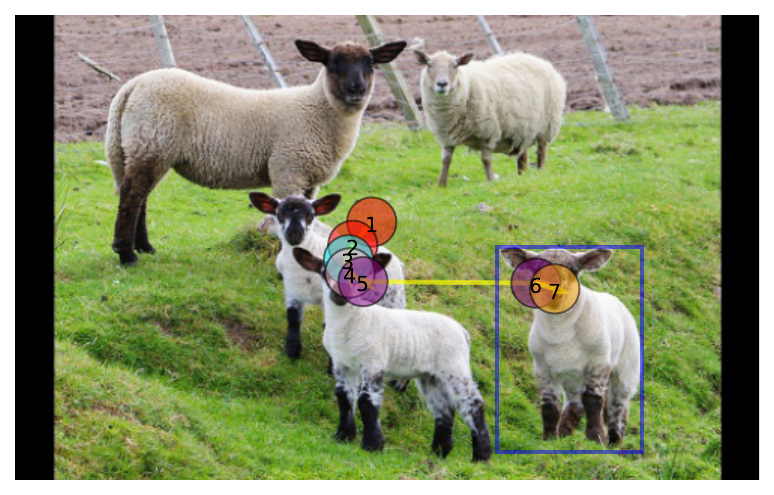} 
 \includegraphics[width=\subFigSz]{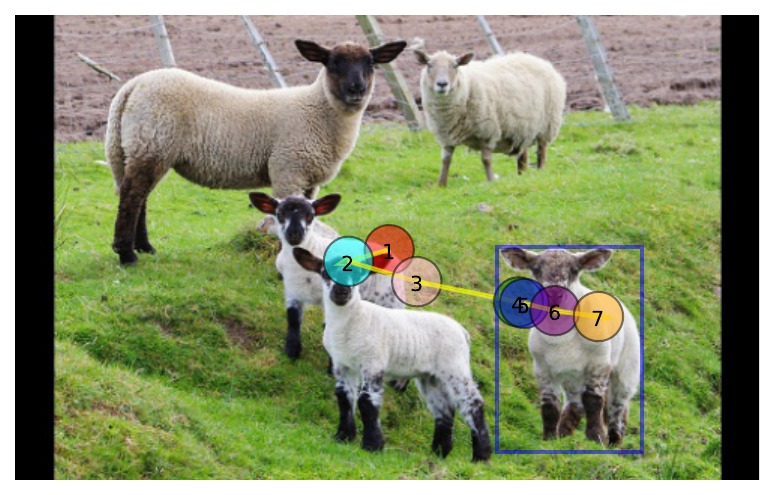}  
  \includegraphics[width=\subFigSz]{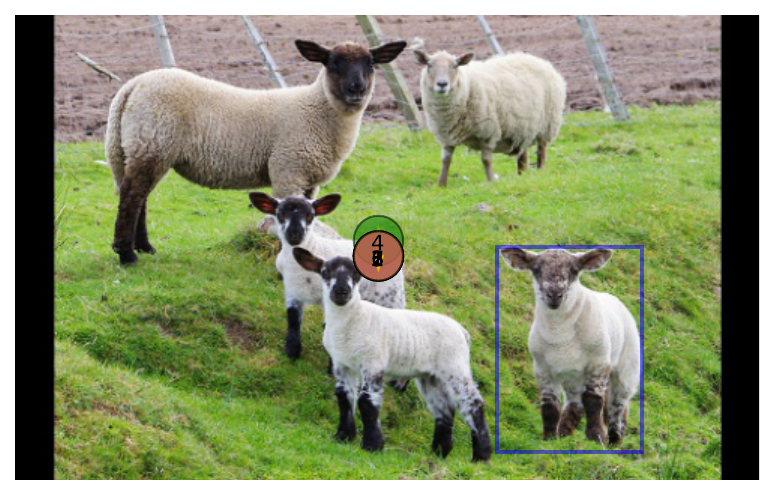} 
  \includegraphics[width=\subFigSz]{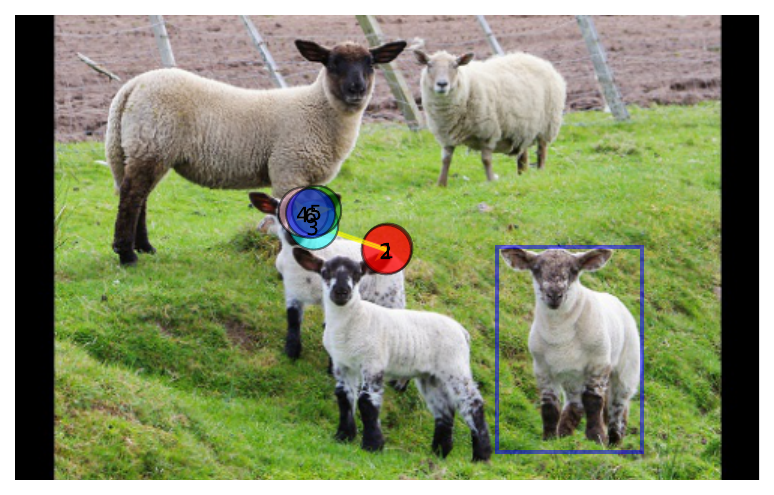} 
  \makebox[\subFigSz]{\small{Human}}
\makebox[\subFigSz]{\small{ART (proposed)}}
\makebox[\subFigSz]{\small{Chen \etal~\cite{chen2021predicting}}}
\makebox[\subFigSz]{\small{Gazeformer-ref~\cite{mondal2023gazeformer}}}\\
  \includegraphics{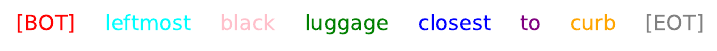} \\
  \includegraphics[width=\subFigSz]{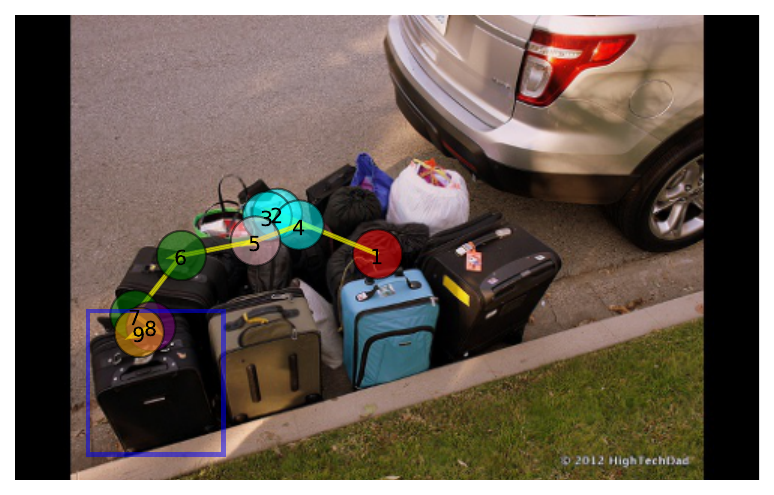} 
 \includegraphics[width=\subFigSz]{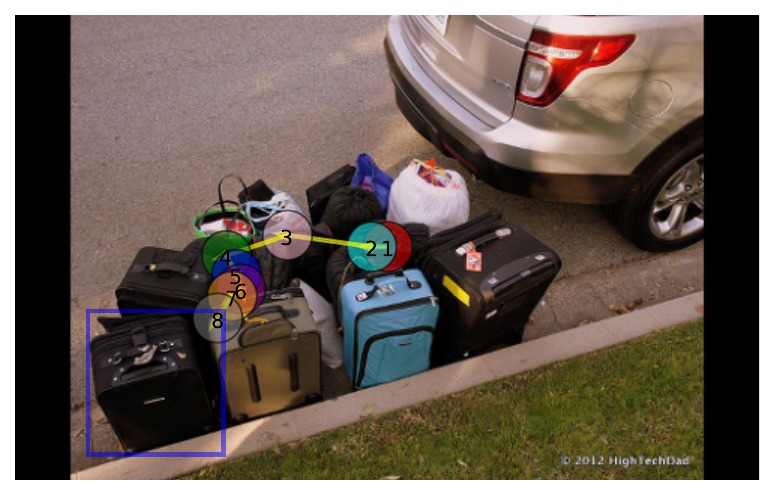}  
  \includegraphics[width=\subFigSz]{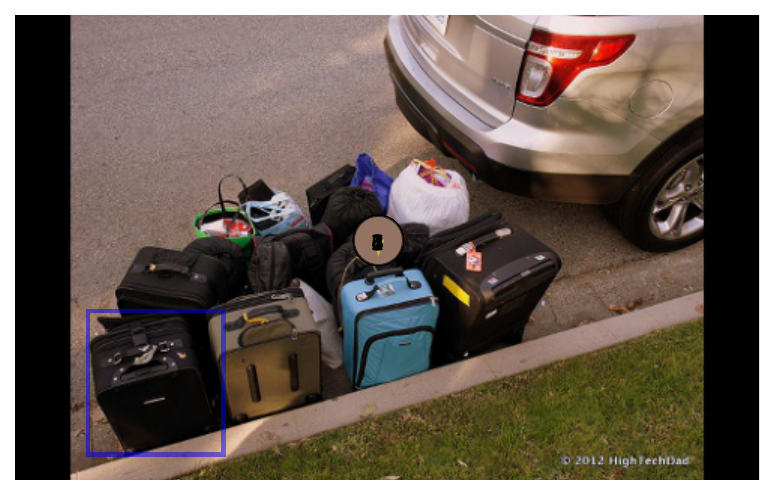} 
  \includegraphics[width=\subFigSz]{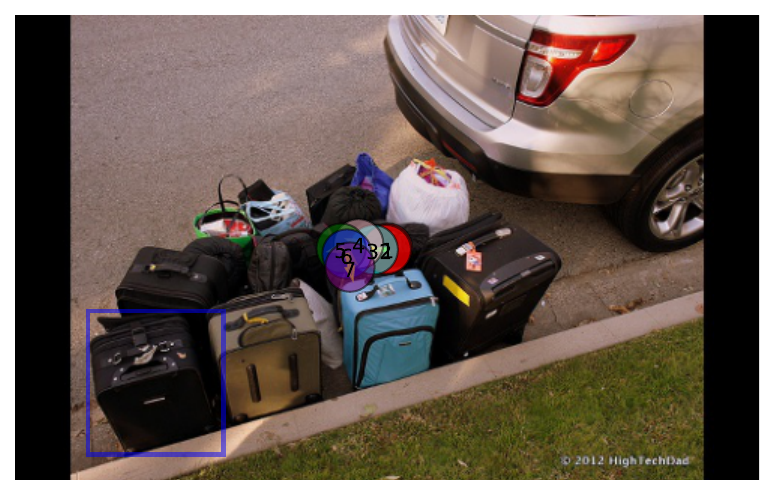} \makebox[\subFigSz]{\small{Human}}
\makebox[\subFigSz]{\small{ART (proposed)}}
\makebox[\subFigSz]{\small{Chen \etal~\cite{chen2021predicting}}}
\makebox[\subFigSz]{\small{Gazeformer-ref~\cite{mondal2023gazeformer}}}\\
  

  \includegraphics{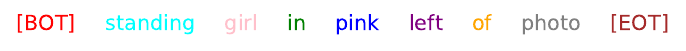} \\
  \includegraphics[width=\subFigSz]{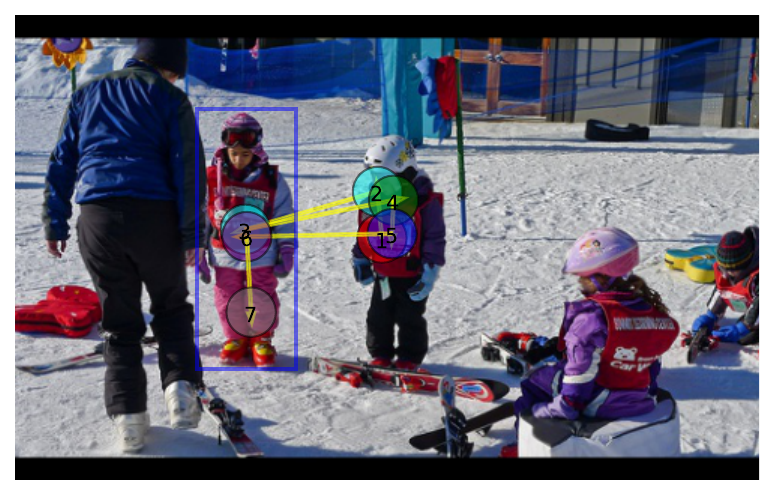} 
 \includegraphics[width=\subFigSz]{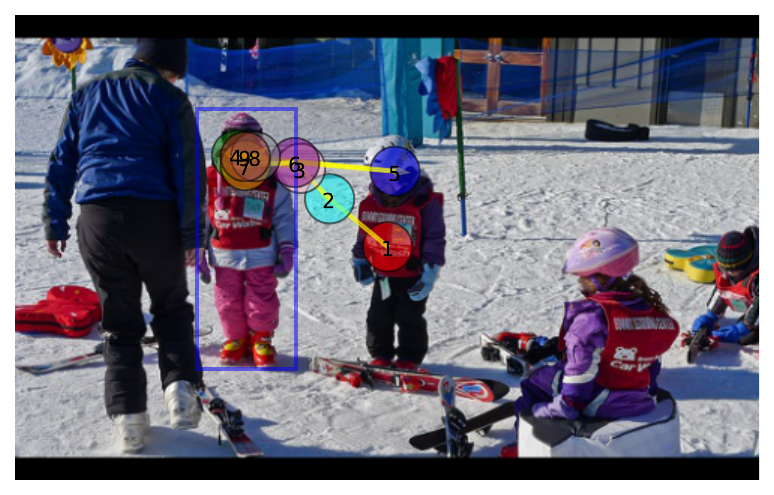}  
  \includegraphics[width=\subFigSz]{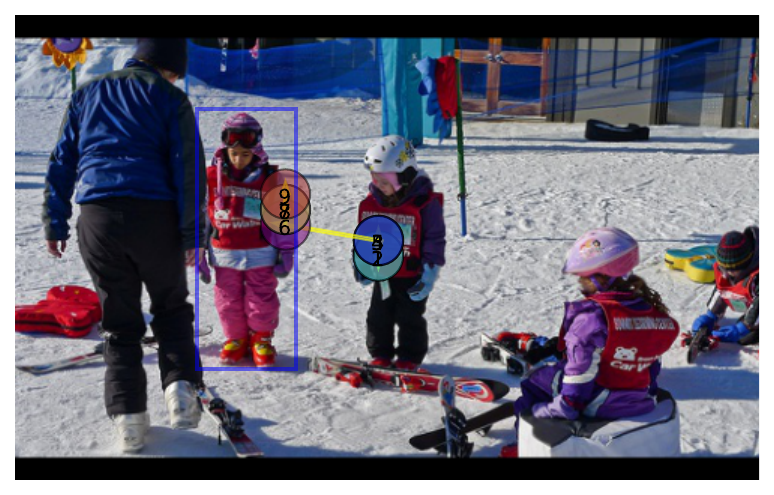} 
  \includegraphics[width=\subFigSz]{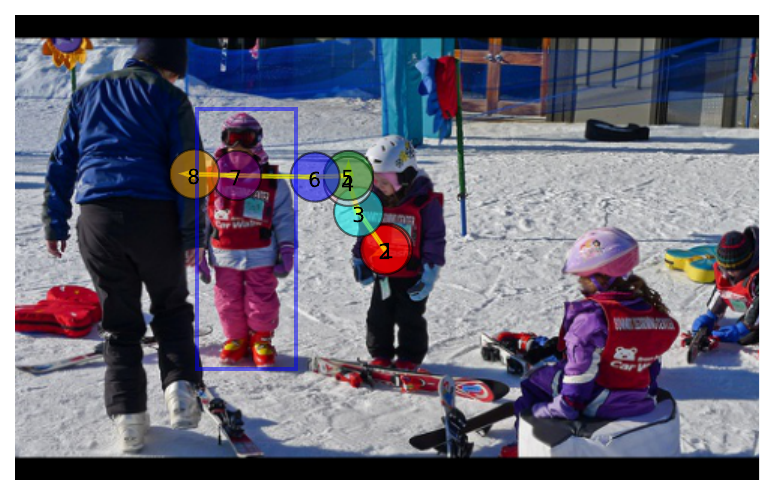} 
  \makebox[\subFigSz]{\small{Human}}
\makebox[\subFigSz]{\small{ART (proposed)}}
\makebox[\subFigSz]{\small{Chen \etal~\cite{chen2021predicting}}}
\makebox[\subFigSz]{\small{Gazeformer-ref~\cite{mondal2023gazeformer}}}\\
\caption{{\bf Qualitative results.}  Scanpaths from humans and three scanpath prediction models on three trials exhibiting strategic fixation behavior. Fixations (denoted by circles numbered with fixation order) are color-coded to corresponding words in the referring expression (above each row). Fixations color-coded to \texttt{[BOT]} occurred before the expression started, and those color-coded to \texttt{[EOT]} occurred after the expression ended. Blue bounding boxes indicating the referred objects are not visible during trials. Our model generates the most human-like scanpaths for incremental object referral.}
\label{fig:qualitative_results}
\end{figure}

\subsection{Results}

We train ART and the baselines on the RefCOCO-Gaze training set and evaluate them on the test set. Results are in \Tref{table:main_results}. ART outperforms baselines on all metrics by significant margins. 
We hypothesize that ART performs best because it \textit{includes} an object referral model and jointly trains on grounding and gaze prediction objectives. In contrast,  Chen \etal~\cite{chen2021predicting}~use a frozen MDETR model trained only on complete RefCOCO expressions, and the Gazeformer variants (Gazeformer-ref and Gazeformer-cat) lack grounding subnetworks to train on object grounding. Hence, both baselines are unable to learn the auxiliary grounding tasks on partial expressions. Interestingly, despite using no spatial and attribute information, Gazeformer-cat is almost as predictive as Gazeformer-ref, underscoring the importance of target category estimation. Note that these analyses focused on spatial attention, but in the supplement, we show that ART also outperforms baselines in terms of fixation duration prediction as well. We also show in the supplement that ART \textit{generalizes well to categorical search} when trained and evaluated using COCO-Search18~\cite{chen2021coco} dataset. 

\Fref{fig:qualitative_results} shows qualitative results comparing the sequences of fixations from ART and the baselines to the behavioral data. In these three examples, ART finds the referred object and generates efficient, human-like scanpaths. ART also exhibits several strategic fixation patterns that we observe in the human data. 
In the top row, ART \textit{waits} near the center until after getting the word ``right'', which conveys information about the referred sheep. A \textit{scanning} gaze pattern appears in the second row, where both the person and ART scan multiple bags, thereby enabling the correct one to be located when the disambiguating information arrives at the end of the expression. The third row exemplifies \textit{verification}. ART successfully finds the correct target on fixation~\#3 after input of the word ``girl'', but then makes another fixation (\#5) to the girl in the center after getting the word ``pink'', presumably to verify which of the girls is pinker before returning to the one on the left on the next fixation (\#6). 


\subsection{Ablation Studies}
\setlength{\tabcolsep}{3pt}
\begin{table}[tb]
\centering
\caption{{\bf Ablation studies on ART model.} If either $\mathcal{L}_{bbox}$ or $\mathcal{L}_{target}$ is included, the loss is applied in \textit{both} pre-training and training phases.} 
\label{table:ablate_results}
\begin{tabular}{c|ccc|ccc}
\toprule 
Ablation \# & Pre-training & $\mathcal{L}_{bbox}$ & $\mathcal{L}_{target}$ & $SS\bm{\uparrow}$ & $SS_{pack}\bm{\uparrow}$&$CC_{pack}\bm{\uparrow}$\\ \midrule 

1 &  $\times$ & $\times$ & $\times$ & 0.309 &  0.257 &  0.222 \\
2 &  $\checkmark$ & $\checkmark$ & $\times$ &  0.321 &  0.279 &  0.239 \\
3 & $\checkmark$ & $\times$ & $\checkmark$ & 0.292  & 0.260  & 0.216\\
4 & $\times$& $\checkmark$  & $\checkmark$ & 0.304  & 0.257 & 0.215\\
5 & $\checkmark$& $\checkmark$  & $\checkmark$ & \textbf{0.359}  & \textbf{0.292} & \textbf{0.280}\\
\bottomrule
\end{tabular}
\end{table}

We performed a number of ablations (in \Tref{table:ablate_results}) on ART to probe the effects of pre-training and inclusion of grounding losses on its performance, which we evaluate using the RefCOCO-Gaze test set. As evidenced by comparing Ablations 4 and 5, pre-training significantly improves performance. We also observe that without pre-training, the grounding losses $\mathcal{L}_{bbox}$ and $\mathcal{L}_{target}$ slightly degrade performance (compare Ablations 1 and 4). When trained from scratch on the small gaze dataset, these auxiliary tasks introduce noise to the optimization of the gaze prediction objective using $\mathcal{L}_{gaze}$. Including one of $\mathcal{L}_{bbox}$ (Ablation 2) and $\mathcal{L}_{target}$ (Ablation 3) losses in pre-training and training does not improve performance significantly (although $\mathcal{L}_{bbox}$ seems to be more beneficial than $\mathcal{L}_{target}$), whereas including \textit{both} $\mathcal{L}_{bbox}$ and $\mathcal{L}_{target}$ (main model and Ablation 5) yields the best performance. This demonstrates that both object localization and target category estimation tasks are integral to the object referral process. 
We note that even when ART is not pre-trained on RefCOCO (Ablations 1, 4), it still outperforms baselines like the two Gazeformer variants that are also not pre-trained on RefCOCO. 
See the supplement for more metrics, ablations and analyses.

\section{Conclusions and Discussion}
\label{sec:conclusion}
How do humans integrate vision and language information to guide their attention to target goals? To study this question we introduced the \textit{incremental object referral} task, a naturalistic version of an object referral task in which people must incrementally integrate the visual information that they are actively collecting with each location that they fixate in the image, with the language information that they are hearing about the target object. Our task therefore provides an experimental context for studying how humans use the spoken language of another to dynamically control the visual information that they sample from the world. 
We also introduced a model that we call \textit{ART} that similarly generates sequences of gaze fixations that occur as this vision-language integration is happening. ART has a multimodal transformer architecture that it uses to learn how to incrementally generate \textit{packs} of fixations for each word in the referring expression. 
To provide the human behavior needed to train ART, we collected a high-quality and large-scale dataset called \textit{RefCOCO-Gaze}.
We trained and evaluated ART and several competitive baselines on RefCOCO-Gaze and found that ART outperformed other baselines by significant margins on multiple metrics. We also performed extensive ablation analyses to show how pre-training ART on RefCOCO, and the addition of auxiliary grounding losses, significantly contributed to its superior performance. 
Qualitative analysis revealed that ART showed human-like effects of visual and linguistic target ambiguity on its attention behavior through higher-level strategic forms of integrating vision and language information, expressed as distinct \textit{waiting}, \textit{verification}, and \textit{scanning} strategies. We believe that ART will be instrumental for predicting gaze in time-sensitive, voice-assisted HCI applications (especially AR/VR) where predicting future eye movements will enable seamless human-computer interactions.

A current limitation of ART is that it treats the referring expression as text and not audio, thereby ignoring the phonological factors influencing vision-language disambiguation and attention control. Future work will explore representing these phonological factors as well. 

\myheading{Acknowledgements.} {\small  This project was supported by US National Science Foundation Award IIS-1763981, IIS-2123920, DUE-2055406, and the SUNY2020 Infrastructure Transportation Security Center, and a gift from Adobe.}

%
%
\clearpage
\chapter*{Supplementary Material}
\setcounter{section}{0}
\renewcommand{\thesection}{\Roman{section}}

In the supplementary material provided below, we present additional experiments, visualizations and details of our work on gaze prediction during \textit{incremental object referral} task using the \textit{Attention in Referral Transformer (ART)} model and \textit{RefCOCO-Gaze} dataset. The specific sections are as follows:

\begin{itemize}
    \item We provide details about how we selected the stimuli, i.e. the images and referring expressions, from RefCOCO dataset to create RefCOCO-Gaze dataset. (Section~\ref{sup:referral-selection}).
    \item We discuss the gaze recording method we used to collect human fixations for RefCOCO-Gaze dataset along with analysis of the collected gaze data (Section~\ref{sup:gaze-methods}).
    \item We provide a comparative analysis of our proposed RefCOCO-Gaze dataset and other related gaze datasets discussed in the main text (Section~\ref{sec:refcocogaze_comp}).
    \item We provide details of various components of the ART model along with the pre-training and training procedures (Section~\ref{sec:art_details}).
     \item We provide implementation details of several scanpath metrics used for evaluation in the main text (Section~\ref{sec:metrics_supp}).
    \item We augment ART with fixation duration prediction capability and report the experimental results for ART model and other baselines on RefCOCO-Gaze with respect to both fixation location and fixation duration prediction (Section~\ref{sec:duration_predictions}).
    \item We show that ART generalizes to categorical search task when trained and evaluated using COCO-Search18~\cite{chen2021coco} dataset (Section~\ref{sec:cocosearch}).
    \item We provide implementation details of the baselines - Random Scanpath, OFA~\cite{wang2022ofa}, Chen~\etal~\cite{chen2021predicting}, Gazeformer-ref~\cite{mondal2023gazeformer}, and Gazeformer-cat~\cite{mondal2023gazeformer} (Section~\ref{sec:baselines_supp}).
       \item We augment the experimental results for the ablation studies on ART model, which are discussed in the main text, with additional metrics (Section~\ref{sec:ablations_comp}).
    \item We present additional ablation studies investigating the effects that the components of ART model have on performance and also include additional analysis on the ablations (Section~\ref{sec:ablations_supp}).
    \item We explore the effects of next word token prediction task on the performance of ART model (Section~\ref{sec:next_word_supp}).
    \item We present additional qualitative examples of scanpaths generated by human participants, our ART model and 
 other competitive baseline models. (Section~\ref{sec:qual_supp}).
    \end{itemize}

\section{Image and Referring Expression Selection Details}
\label{sup:referral-selection}

\begin{figure}[ht!]
  \centering

  \includegraphics[width=1.0\textwidth]{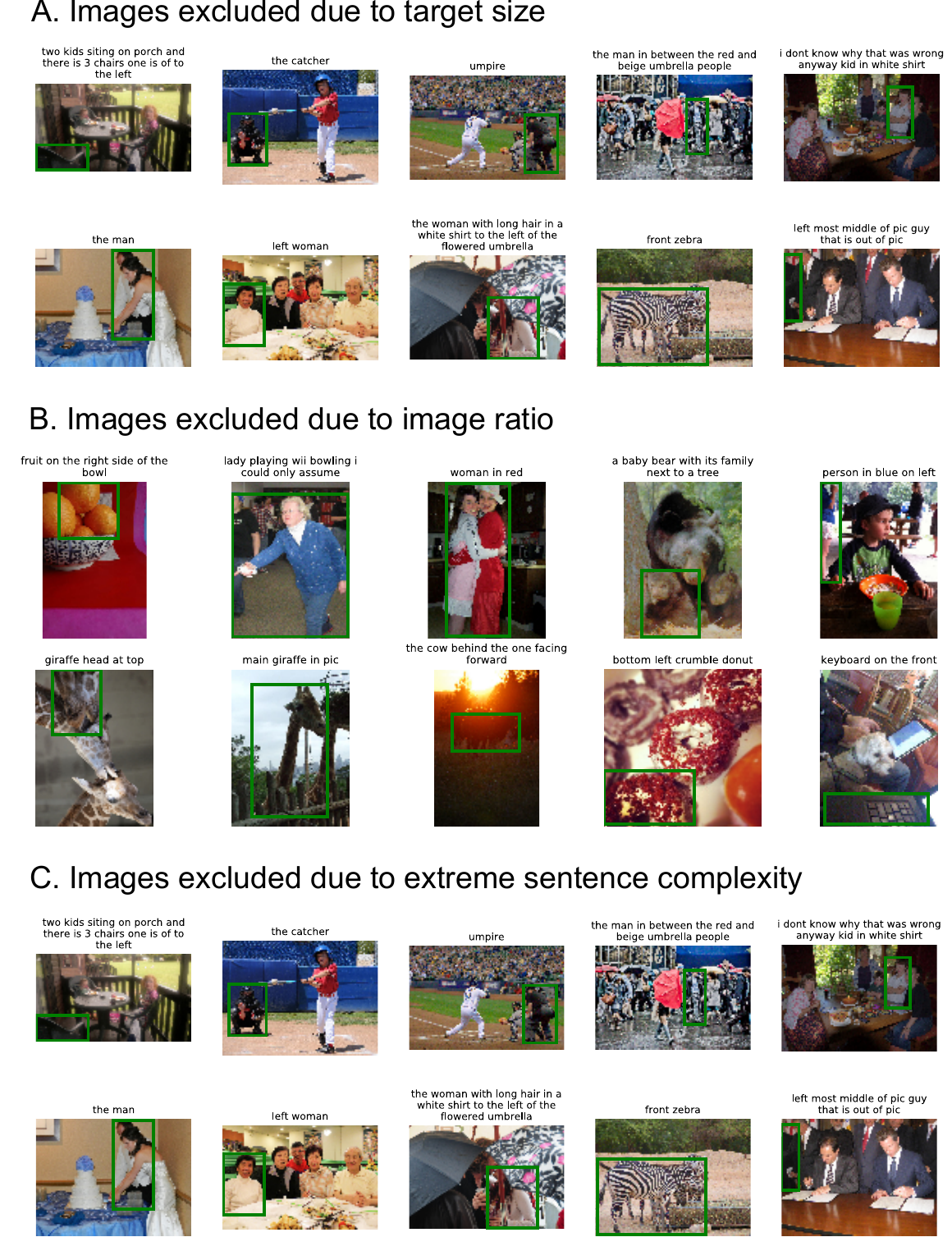}
    \caption{Excluded Samples}
  \label{fig:excluded-examples}
\end{figure}

We utilized a subset of the RefCOCO dataset~\cite{yu2016modeling} (the original UNC split) to create our dataset. RefCOCO dataset consists of referring expressions collected for 50,000 target objects present in 19,994 COCO~\cite{linMicrosoftCoco2014} training images. RefCOCO was carefully curated such that each image contained at least two objects of the same object category as the target object. To ensure the reliability of our gaze data, we selected the longest referring expression amongst the multiple referring expressions collected for each target object. To eliminate stimuli that might produce inaccurate gaze patterns due to low quality or extreme difficulty, we further refined our dataset by excluding images and referring expressions that did not meet the following criteria. For detailed examples of such exclusions, please refer to Fig.~\ref{fig:excluded-examples}. 


\begin{itemize}

  \item \textbf{Target Size}: We excluded data where the size of the target object, as measured by the area of its bounding box, was larger than 10\% of the total image area. 

  \item \textbf{Image Ratio}: We excluded data with images whose width-to-height ratios were outside the range of 1.2-2.0 (based on a screen ratio of 1.6). We did this to eliminate very elongated images, which might distort normal viewing behavior. 

  \item \textbf{Sentence Complexity}: We excluded data where the referring expression sentence was either too simple or too complex. We measured sentence complexity using the metric introduced by Liu et al.~\cite{liu2020norm}, which is correlated with sentence length and frequency. Specifically, we excluded data where the referring expression language complexity was below the 10th percentile (e.g., "the girl") or above the 90th percentile (e.g., "second row from left to right second one up from bottom..."). This ensured that the length of the remaining referring expressions ranged from 2 to 10 words, with a median of 4. The original dataset had a wider range of sentence lengths, from 1 to 39 words, with a median of 4.
  

  
\end{itemize}

After applying the aforementioned exclusion criteria, we were left with 7,568 referring expressions from 72 categories. However, this number exceeded our resources for gaze data collection. Therefore, we decided to further trim the dataset while maintaining a balanced distribution of target categories. To do so, we removed entire categories if the application of the exclusion criteria left fewer than 100 referring expressions per category. Then, we randomly selected up to 150 data points per category. This process resulted in a dataset of 2,422 referring expressions from 18 categories. Finally, we conducted a manual exclusion process to remove any referring expressions or images containing obscene, inappropriate, or irrelevant content (e.g., blood, nudity, slang). We also manually removed any data points with spelling mistakes, or incorrect or poor target descriptions. In total, 328 data points were removed during this manual exclusion process, yielding 2,094 image-expression pairs for the dataset.

\begin{figure}[h]
  \centering
  
  \includegraphics[width=1.0\textwidth]{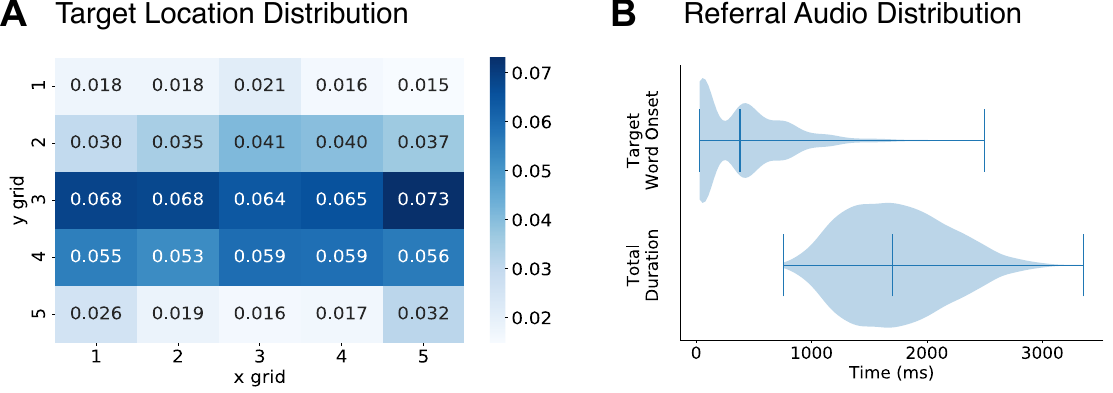}
  \caption{Stimuli Statistics. A: Spatial distribution of target locations B: Temporal distribution of target word onset}
  \label{fig:stimuli}
\end{figure}

\begin{figure}[t]
  \centering
  \includegraphics[width=0.9\textwidth]{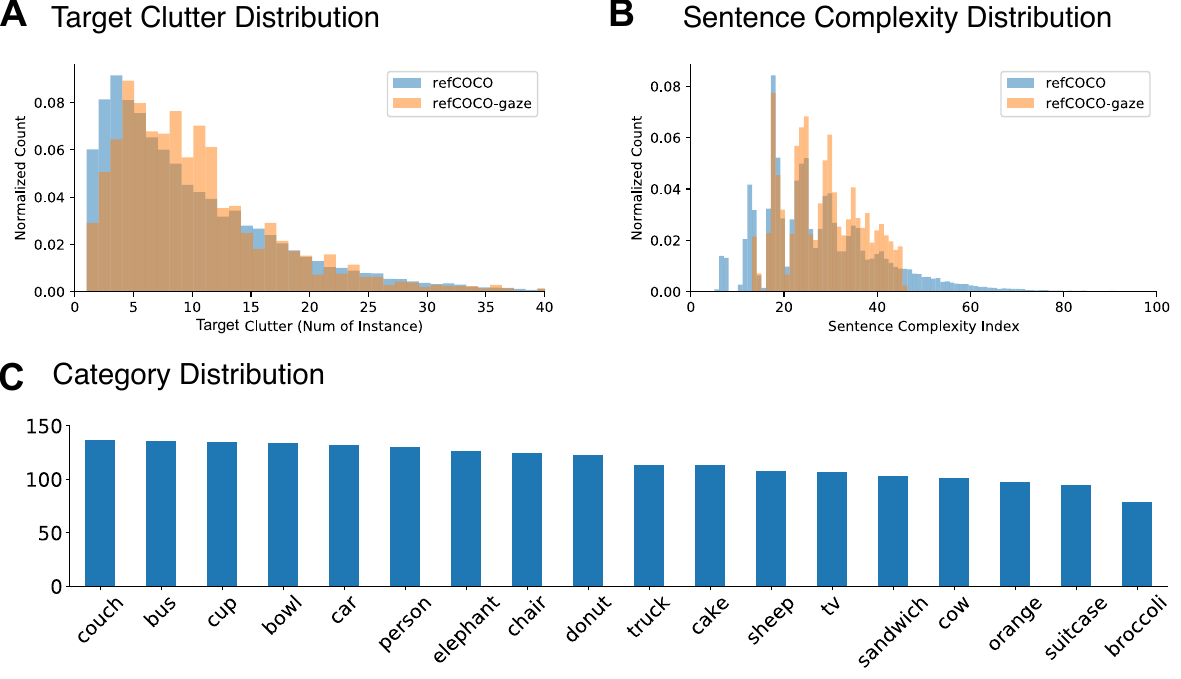}
  \caption{Descriptive statistics for RefCOCO-Gaze images and referral expressions.}
  \label{fig:statistics}
\end{figure}

\section{Gaze Recording Methods}
\label{sup:gaze-methods}

\subsection{Participants}
Our dataset was collected from 220 participants, consisting of 155 male, 63 female, and 2 non-binary individuals. Participants were undergraduate students from our institution who were recruited for extra credit in a psychology class and had normal or corrected-to-normal vision. The age range of participants was between 18 and 33 years. Among the participants, 28 were non-native English speakers but rated their fluency level as either very good or good. This study had Institutional Review Board (IRB) approval.

\subsection{Stimuli}
\label{sec:stimuli}
Images were resized and padded to fit to the computer screen size and resolution (1680$\times$1050 pixel resolution). Fig.~\ref{fig:stimuli}A displays the spatial distribution of target locations in the images, which are evenly distributed but with slightly higher probability around the center-bottom area. 
Spoken referring expression were generated using Google Text to Speech API\footnote{https://pypi.org/project/google-cloud-texttospeech/} commonly known as the gTTS available in Python. Fig.~\ref{fig:stimuli}B shows the temporal distribution of target word onset, which was measured by the timing of the target word from generated audio. The original dataset did not provide the target word (i.e., the word in the sentence that refers to the target object). Therefore, we manually annotated the target word for each referring expression using the consensus of two annotators. Example target words are provided in Table~\ref{sup:target-names}. 
The target word is usually referred to at the beginning of the sentence (median time of 0.4 seconds), and the total duration of the audio mostly ranges from 1 to 3 seconds, with a median of 1.7 seconds. 

Our dataset closely matches the distribution of the original RefCOCO dataset (panels A and B of Fig.~\ref{fig:statistics}), capturing the distributions of target clutter (i.e., the number of instances of the same target category in the image) and sentence complexity (as measured by Liu \etal~\cite{liu2020norm}). Fig.~\ref{fig:statistics}C shows the category distribution of target objects in the dataset. These categories were well-balanced and span a wide semantic range, from animate objects (e.g., person, sheep) and indoor objects (e.g., chair, cup) to outdoor vehicles (e.g., car, truck). 

\begin{table}[ht!]
    \centering
    \caption{Target Names}
    \label{sup:target-names}
    \begin{tabularx}{\textwidth}{@{}lX@{}}
    \toprule
    Category &
      Target Names Used for Referral \\ \midrule
      \endhead
    person &
      coach, figure, pic, jeans, player, girl, arm, row, thingy, shirt, girls, kid, hat, red, skier, picture, head, woman, left bottom, hand, slider, child, lady, man, boy, catcher, jacket, person, red and black, green, guy \\
    elephant &
      rump and tail, ear, elephant butt, butt, one, camera, animal, elephant, baby, corner, legs \\
    sheep &
      sheep, goat or sheep, cow, area, animal butt, animals face, leg, ship, one, lamb, sheep butt, animal, guy, face, goat, calf \\
    cow &
      brown, cow, leg, camel, one, band, animal, bull, cows, corner, critter, legs, goat, calf \\
    bus &
      trolley, double decker, van, phone, bus, ride, deck, decker, glass, train, truck, vehicle, thing, rectangle, car \\
    car &
      taxi, benz, van, reflection, area, car, screen, mirror, police, suv, truck, vehicle, cab, ford, black \\
    truck &
      hummer, area, van, semi, suv, door, car, vehicle, firetruck, bus, fedex, machine, thing, part, truck, item, fire truck, corner, rig, tarp, trailer \\
    couch &
      loveseat, armchair, frame, seat, corner, ottoman, chair, love seat, thingy, couch, orange, seat cushion, pillows, leg rest, cushion, table, furniture, footstool, chairs, sofa, bed, thing, pillow \\
    chair &
      woven, center, thing, chairs, tray, object, couch, jacket, item, bench, pattern, corner, chair, seat, lady, seat cushion \\
    tv &
      laptop, monitor, tablet, tv, poster, screen, tv screen, bruce lee, face, computer, monitor screen, desktop, girl, sign, spot, computer screen \\
    suitcase &
      case, box, container, briefcase, luggage, area, space, bag, black, item, suitcase, corner, chair, thing, trunk \\
    bowl &
      right, bananas, cup and spoon, pot, dish, corner, thing, row, container, cup, things, section, bowl, butter, sauce, pan, kiwi, plate, food, left bowl, grapes, tuna, chips, broccoli, pottery piece, hot dog, fruit slices, soup, stuff, apples, dip \\
    cup &
      tea, pot, whatever, one, juice, second, frosty, dish, candle, milk, blender, container, cup, section, jar, mug, beer, coffee, coke, drink, pitcher, toothbrush, glass, water, thing, stuff, bottle \\
    donut &
      plate, sprinkles, item, food stuff, donuts, food, cheerio, donut, chocolate ice, bun, pastry, dessert, corner, skewer, doughnut, thing, striped, row \\
    cake &
      muffin, pie, one, pastry, corner, thing, ice cream, row, item, pile, orange, hat, roll, plate, umbrella, food, dessert, brownie, cupcake, cake, fruit, cookie, frosting, chocolate, bread, center, biscuit, train car, cake slice \\
    sandwich &
      taco, waffle, burger, ball, pastry, wrap, sammy, toast, flower, sub, bowl, palte, roll, plate, sandwich, food, appetizer, bun, half, banana slices, bread, meat, thing, piece \\
    orange &
      one, slice, apple, left, thing, orange slice, bowl, oranges, pieces, lime, grapefruit, fruit, food, lemons, front, orange, lemon, stem, egg, row \\
    broccoli &
      broccoli piece, broccoli pieces, greens, spinach, food, blur, veggie, patch, thing, piece, green, goop, basket, piece of broccoli, broccoli \\ \bottomrule
    \end{tabularx}
    
\end{table}

\clearpage
\subsection{Procedure and Apparatus}
Gaze was recorded using the EyeLink 1000 eye-tracker (SR Research Ltd., Ottawa, Ontario, Canada) and the data were exported using the EyeLink Data Viewer software package (also from SR Research Ltd.). During the experiment, the presentation of images was controlled using Experiment Builder software (SR Research Ltd., Ottawa, Ontario, Canada). The stimuli were displayed on a 22-inch LCD monitor, positioned at a viewing distance of 47cm from the participant, with the help of chin and head rests. This resulted in a horizontal and vertical visual angle of 54$^\circ$ $\times$ 35$^\circ$, respectively. At the beginning of each trial, participants were instructed to fixate on a central point but were free to move their eyes while searching for the target. Eye movements were recorded throughout the experiment using the EyeLink 1000 eye-tracker in tower-mount configuration. Prior to each block or whenever necessary, the eye-tracker was calibrated using a 9-point calibration method, and the calibration was not accepted unless the average calibration error was below 1.0$^\circ$ and the maximum error was below 1.5$^\circ$. The experiment was conducted in a quiet laboratory room under dim lighting conditions. All responses were recorded using Microsoft Game controller triggers. The following instructions were provided to the participants prior to the gaze data collection process: \\

\textit{``We wish to observe your natural eye-movement behavior while searching for a referred target. You will be shown 100 images with spoken referring expressions describing the target’s location and appearance. Your job is to find a target AS QUICKLY AND ACCURATELY AS POSSIBLE. When you find a target, please press any button on the top side of the controller. We will analyze your gaze later and measure accuracy by checking whether your gaze land on the target correctly at the time you press. So please make sure you press the button WHILE you are looking at the target. Please press the button as soon as you find the referred target. You can browse each image up to 5 seconds after the sound ends. There will be a break around halfway through the experiment, but if you need an additional break during the experiment, let the experimenter know anytime.''}

\subsection{Preprocessing}
Fixations were detected from raw gaze samples using the EyeLink online parser, which applied velocity and acceleration thresholds of 30$^\circ/s$ and 8000$^\circ/s^2$, respectively. Fixations with a duration lower than 60ms were filtered out, but all other fixations were retained. The initial raw dataset consisted of 21,898 valid scanpaths. However, to ensure data reliability, we removed trials where participants did not find the target within a given time limit of 5 seconds or reported not finding the target in the survey. We also eliminated trials where any of the participant's fixations did not land within the target bounding box, resulting in the removal of 10\% of the entire dataset and leaving us with 19,738 scanpaths. Additionally, we observed that 6\% of trials had the participant's final fixation not within the target area, which may have been due to them moving away from the target as they pressed the button. To address this, we trimmed the fixations up to the last fixation that landed on the target, thereby ensuring that only fixations relevant to the target search were included in the analysis.

\subsection{Gaze Data Analysis for RefCOCO-Gaze}

\myheading{RefCOCO-Gaze fixations are intention-driven.} As can be seen by comparing the top two rows in Table~\ref{table:main_results} from the main text, the inter-observer agreement metrics (row 1, labeled ``Human'') far exceed the Random baseline metrics (row 2). Based on this observation, and findings of previous behavioral work~\cite{torralba2006contextual, xu2017has} suggesting that high inter-observer agreement mark similar task-driven attention allocation across individuals, we infer that the fixations in RefCOCO-Gaze are not random but rather, intention-driven and under attention control.

\myheading{Target Localization analysis.} Analysis of the gaze data collected in our incremental object referral task revealed that on 9.76\% of the trials, participants failed to either fixate on the correct target or to localize the target within the 5 second limit. Mean saccade amplitudes for successful and failed localizations were 192.948 (standard deviation=75.86), and 191.662 (standard deviation=78.94), respectively. On the other hand, mean fixation durations (in msecs) for successful and failure localizations were 280.741 (standard deviation=114.38) and 277.432 (standard deviation=126.68), respectively. As is evident, the gaze statistics of average saccade amplitude and average fixation duration did not significantly differ between the successful localizations and failed localizations (T-test revealed p-value$=$0.437 for average saccade amplitude, and p-value$=$0.247 for average fixation duration -- both not statistically significant since p-value$>$0.05). However failure cases yielded statistically significant longer scanpaths (average of 10 fixations for failure, 8 for success; p-value$<$0.05). Failure cases also showed strong positive correlations with scene complexity (Pearson correlation coefficient $r$=0.81) in terms of object instance count in a scene, and referral language perplexity (Pearson correlation coefficient $r$=0.76). These complexity scores tend to be higher for failure cases than for successful ones, suggesting that search performance decreases with increasing scene complexity and linguistic complexity of the referring expressions. 

We also note that in 12.52\% of the trials, observers fixated on the target during exposure time. Yet, for these trials, search ended after a median of 5 words, implying that observers required ample description for confident localization.

\newpage
\section{Comparison of RefCOCO-Gaze with other gaze datasets}
\label{sec:refcocogaze_comp}
Here, we compare related datasets discussed in Related Work section (Section 2 in the main text) in the table below. Our proposed dataset, RefCOCO-Gaze is the \textit{only} gaze scanpath dataset for the incremental object referral task.

{\scriptsize
\begin{tabularx}{\textwidth}{X*{8}{>{\centering\arraybackslash}X}} 
\toprule
Dataset & Apparatus & Task & Gaze recorded [during/after] task description & Stimuli & No. of scanpaths & No. of Subjects & Relevant w.r.t. Object Referral & Relevant w.r.t. Incremental Prediction  \\ \midrule
  \endhead\\
COCO-Search18~\cite{chen2021coco}&Eye-tracker&Categorical Visual Search&After&Images&299037&10&No&No\\\\
\hline\\
AiR~\cite{chen2020air}&Eye-tracker&VQA&After&Images&13173&20&No&No\\\\
\hline\\
Localized Narratives~\cite{pont2020connecting}&Mouse proxy&Image Captioning&During&Images&848749&156&No&\textbf{Yes}\\\\
\hline\\
He~\etal~\cite{he2019human}&Eye-tracker&Image Captioning&During&Images&14000&16&No&\textbf{Yes}\\\\
\hline\\
SNAG~\cite{vaidyanathan2018snag, vaidyanathan2020computational}&Eye-tracker&Image Description&During&Images&3000&30&No&\textbf{Yes}\\\\
\hline\\
OR~\cite{vasudevan2018object}&Face videos&Object Referral&After&Videos&30000&20&\textbf{Yes}&No\\\\
\hline\\
Zhang~\etal~\cite{zhang2022gaze} & Gaze Following & Object Referral & After & Images & $- ^{*}$ & - & \textbf{Yes} & No\\\\
\hline\\
\textit{RefCOCO-Gaze(ours)}&Eye-tracker&Incremental Object Referral&During&Images&19738&220&\textbf{Yes}&\textbf{Yes}\\
\bottomrule
\end{tabularx}
${*}$~Zhang~\etal~\cite{zhang2022gaze} collect 40000 \textit{static gaze heatmaps}, not \textit{spatiotemporal gaze scanpaths}
}

\section{Additional Details of ART}
\label{sec:art_details}

In this section, we share additional details about the ART model, such as details of implementation, architectural design, and hyperparameter choices for our experiments on RefCOCO-Gaze. 

\subsection{Visual Encoder and Language Encoder}

For designing the visual encoder, we use an ImageNet~\cite{deng2009imagenet} pre-trained ResNet-50~\cite{he2016deep} backbone followed by a transformer encoder consisting of 6 standard transformer encoder layers~\cite{vaswaniAttention} with hidden size $d_{vis}=256$ and 8 attention heads. A dropout of 0.1 was applied to the transformer encoder layers. The output of the visual encoder is patch embedding tensor $g_{vis}\in \mathbb{R}^{d_{vis}{\times}hw}$, corresponding to $h\times w$ grid, where $h=10, w=18$. For the language encoder, we use the RoBERTa-base variant~\cite{liu2019roberta} which generates embeddings of dimension $d_{lang} = 768$ for each token in the tokenized text string.  The hyperparameter $l_{lang}$ is set to 32. RoBERTa encodes text tokenized using a Byte-Pair Encoding (BPE)~\cite{sennrich2016neural}. RoBERTa is pre-trained on a large corpus of English data (which includes the BookCorpus~\cite{zhu2015aligning}, English Wikipedia data, the English portion of CommonCrawl News
dataset~\cite{nagel2016} called CC-News, OpenWebText~\cite{Gokaslan2019OpenWeb} and STORIES~\cite{trinh2018simple}) using a Masked Language Modeling (MLM) objective with a dynamic masking scheme. As mentioned in the main text, we specifically use ResNet-50 and RoBERTa backbones for fair comparison because they form the backbones of our baselines Chen \etal~\cite{chen2021predicting} and Gazeformer~\cite{mondal2023gazeformer} variants. Both visual and language encoders are trainable, and not frozen as in Gazeformer~\cite{mondal2023gazeformer}.

\subsection{Visuo-linguistic Transformer Encoder}

For our experiments on RefCOCO-Gaze, ART's visuo-linguistic encoder consists of 6 standard transformer encoder layers~\cite{vaswaniAttention} with hidden size ($d$) 256 and 8 attention heads each. A dropout of 0.1 was applied to all transformer layers in this module. For the bounding box regression and target category prediction heads, a dropout of 0.3 was applied during the pre-training phase while a dropout of 0.2 was applied during the training phase. To deal with scale variation, we normalize the parameters of of ground truth bounding boxes and consequently apply sigmoid activation to the bounding box regression head. 

\subsection{Pack Decoder \& Fixation Prediction}

For our experiments on RefCOCO-Gaze, ART's pack decoder module consists of 6 transformer decoder layers~\cite{vaswaniAttention} with hidden size($d$) 256 and 8 attention heads. A dropout of 0.2 was applied for all transformer decoder layers in this module. For the fixation prediction heads in the fixation prediction module, a dropout of 0.4 was applied. We choose hyperparameters $L_{\mathcal{P}}$ and $L_{\mathcal{C}}$ to be 6 and 36 respectively. Spatial location estimation was done by regressing parameters (i.e. mean and log-variance) of two separate Gaussian distributions using 4 regression heads (two heads each for the two Gaussian distrbutions - one head for estimating mean and the other head for estimating log-variance) in the fixation prediction module. These Gaussian distributions model the x and y co-ordinates (raw unnormalized pixel co-ordinates) of fixations~\cite{mondal2023gazeformer}. The spatial locations are sampled from the Gaussian distributions using the reparameterization trick~\cite{kingma2013auto}. The range of the predicted unnormalized fixation location (x and y) co-ordinates are the respective image dimensions. We do not involve the pack decoder module and the fixation prediction module in the pre-training stage.

\subsection{Pre-training \& Training}
To deal with scale variation, we normalize the parameters of the ground truth bounding boxes during pre-training and training of the bounding box head. We use AdamW~\cite{loshchilov2018decoupled} optimizers for our pre-training and training phases with weight value 1e-4. During the pre-training process, the visual encoder, the language encoder and the visuo-linguistic encoder are all assigned learning rates of 1e-5. During the training process, the visual encoder and the language encoder are both assigned learning rates of 1e-7 while the visuolinguistic encoder is assigned a learning rate of 1e-5, while the rest of the ART model is assigned a learning rate of 1e-4. We pre-train on the RefCOCO training set for 200 epochs with a batch size of 128 and train on RefCOCO-Gaze training set with a batch size of 64 for a maximum of 200 epochs. Note that the visual, language and visuo-linguistic encoders are trainable (not frozen) during the pre-training stage, and all components of the ART model are trainable (not frozen) during the training stage. We ran our experiments on NVIDIA RTX A5000 GPUs.

\section{Additional Details of Metrics}
\label{sec:metrics_supp}

In this section, we provide additional implementation details of the scanpath metrics.

\myheading{$\textbf{SS}$.} This metric is the sequence score between the ground truth and predicted scanpaths over the \textit{entire} referring expression. Hence, this metric considers \textit{only} the valid 2D fixation locations. 

\myheading{$\textbf{SS}_{pack}$.} We might encounter two edge cases while calculating $SS_{pack}$ - either (1) the predicted pack is a null pack or (2) the ground truth pack is a null pack, with both scenarios resulting in empty strings which hinder direct application of string matching algorithm~\cite{needleman1970general}. We handle the first scenario by duplicating the last fixation of previous non-null predicted pack (initial central fixation point in case there are no previous fixations) and handle the second scenario by duplicating the last fixation of previous non-null ground truth pack (initial central fixation point in case there are no previous fixations) - similar to the process for calculating ScanMatch with duration~\cite{cristino2010scanmatch}. Similarly, we also duplicate the last 2D fixation when one of ground truth scanpath or predicted scanpath has terminated and the other one has not.

\myheading{$\textbf{CC}_{pack}$.} For our implementation of $CC_{pack}$, we add a small $\epsilon=1e^{-9}$ to the ground truth and predicted maps to avoid a divide-by-zero error for cases where either the ground truth map or the prediction map is a zero map due to a null pack. 

\myheading{$\textbf{NSS}_{pack}$.} We disregard cases where either ground truth or predicted pack is a null pack while calculating the average for $NSS_{pack}$.  This is because there is no theoretical upper or lower bound of NSS that can be assigned to scenarios where either one or both of ground truth and predicted packs are null packs (resulting in zero action/saliency maps).

\setlength{\tabcolsep}{3pt}
\begin{table*}[bt]
\centering
\caption{Performance of ART and baselines on RefCOCO-Gaze test set when trained and evaluated on both fixation location and fixation duration prediction tasks.}
\label{table:duration_results}
(a) Duration-agnostic metrics\\
\begin{tabular}{lcccccc}
\toprule 
& $SS\bm{\uparrow}$ & $SS_{pack}\bm{\uparrow}$& $FED\bm{\downarrow}$ & $FED_{pack}\bm{\downarrow}$&$CC_{pack}\bm{\uparrow}$& $NSS_{pack}\bm{\uparrow}$\\  \midrule 
Human & 0.400 & 0.317 & 6.573 & 1.278 & 0.283 & 3.112\\ \midrule
Random &  0.189 & 0.133 &  17.735 & 3.005 & 0.094 & 1.689\\
OFA~\cite{wang2022ofa} &  0.216 & 0.170 &  17.084 & 2.901 & 0.174 & 2.175 \\
Chen \etal~\cite{chen2021predicting} &  0.281 & 0.255 &  6.825 & 1.163 & 0.209 & 1.953\\
Gazeformer-ref~\cite{mondal2023gazeformer} &  0.261 & 0.187 &  6.833 & 1.307 & 0.197 & 2.882\\
Gazeformer-cat~\cite{mondal2023gazeformer} &  0.244 & 0.172 &  7.144 & 1.394 & 0.194 &2.664 \\ 
ART (Proposed) &  \textbf{0.356} & \textbf{0.285} &  \textbf{6.410} & \textbf{1.161} & \textbf{0.281} & \textbf{3.539}\\
\bottomrule
\end{tabular}\\
\vskip 0.2in
(b) Duration-aware metrics\\
\begin{tabular}{lccccc}
\toprule 
& $SS^{(t)}\bm{\uparrow}$ & $SS_{pack}^{(t)}\bm{\uparrow}$& $FED^{(t)}\bm{\downarrow}$ & $FED_{pack}^{(t)}\bm{\downarrow}$ &  $MM_{dur}\bm{\uparrow}$\\  \midrule 
Human & 0.379 & 0.215 & 38.153 & 8.204 & 0.589\\ \midrule
Random & 0.169 & 0.097 & 108.296 & 18.395 & 0.688\\
OFA~\cite{wang2022ofa} & 0.206 & 0.124 & 103.347 & 17.868 & 0.688\\
Chen \etal~\cite{chen2021predicting} &   0.272 & 0.157 & 42.058 & 8.224 & 0.633\\
Gazeformer-ref~\cite{mondal2023gazeformer} & 0.236 & 0.166 & 39.104 & 7.131 & 0.617\\
Gazeformer-cat~\cite{mondal2023gazeformer} &  0.224  & 0.161 & 39.937 & 7.216 & 0.519\\ 
ART (Proposed) &   \textbf{0.332} & \textbf{0.199} & \textbf{35.997} & \textbf{7.120} & \textbf{0.696}\\
\bottomrule
\end{tabular}\\
\end{table*}

\section{Fixation Duration Prediction with ART}
\label{sec:duration_predictions}

ART is also capable of predicting the fixation durations of humans. We model fixation durations as Gaussian distributions, similar to how we model fixation locations. First, we reparameterize a fixation $\mathbf{p}^k_i$ using five parameters: $x$-location $x^k_i$, $y$-location $y^k_i$, \textit{fixation duration} $d^k_i$, the pack number $k$ (i.e., the index of the pack the fixation belongs to), and the within-pack index $i$ (which we call \textit{order}). We then add two fixation duration regression heads (along with the already existing fixation location regression heads) to the fixation prediction module to estimate parameters (i.e., mean and log-variance) of a Gaussian distribution modeling fixation durations.  Fixation durations $d_{i}^{k}$ are sampled from this Gaussian distribution using the reparameterization trick~\cite{kingma2013auto}. Let the predicted pack of fixations $\mP_k = \{(x_{i}^{k}, y_{i}^{k}, d_{i}^{k})\}_{i=1}^{L_{\mathcal{P}}}$, and ground-truth pack of fixations $\hat{\mP}_k = \{(\hat{x}_{i}^{k}, \hat{y}_{i}^{k}, \hat{d}_{i}^{k})\}_{i=1}^{l^{k}}$ where $l^{k}$ is the length of the ground truth pack. Moreover, let $\hat{v}_{i,t}^{k}$ be a binary scalar representing ground truth of the $i^{th}$ token in $\mathcal{P}_k$ belonging to the token class $t\in T \text{ where } T = \{\texttt{FIX}, \texttt{PAD}, \texttt{EOS}\}$. Also let $v_{i,t}^{k}$ be the probability of that token belonging to token class $t$ as estimated by our model. For accomodating the additional fixation duration prediction objective, an $L_1$ loss $\mathcal{L}_{d}^{k}$ between ground-truth fixation durations $\hat{d}_{i}^{k}$ and predicted fixation durations $d_{i}^{k}$ is added to the formulation of $\mathcal{L}_{gaze}$ in Equation 1 of the main paper. Hence, upon accounting for fixation duration prediction along with fixation location prediction, $\mathcal{L}_{gaze}$ for a minibatch of size $M$ now becomes: 

\begin{align}
   & \mathcal{L}_{gaze} = \frac{1}{M} \sum_{k=1}^{M} \left( \mathcal{L}_{xy}^{k} + \mathcal{L}_{token}^{k} + \mathcal{L}_{d}^{k}  \right).
\end{align}

Here $\mathcal{L}_{xy}^{k} = \frac{1}{l^{k}}  \sum_{i=1}^{l^{k}} \left( |x_i^{k}  - \hat{x}_i^{k}| + |y_i^{k}  - \hat{y}_i^{k}|\right)$, $\mathcal{L}_{d}^{k} = \frac{1}{l^{k}}  \sum_{i=1}^{l^{k}} \left( |d_i^{k}  - \hat{d}_i^{k}|\right)$, and $\mathcal{L}_{token}^{k} = - \sum_{i=1}^{L_{\mathcal{P}}} \sum_{t\in T} \hat{v}_{i,t}^{k}\log(v_{i,t}^{k})$. Hence, the total multi-task loss $\mathcal{L}$
that we use to train our ART model for both fixation location prediction and fixation duration prediction is $\mathcal{L} = \mathcal{L}_{gaze} + \mathcal{L}_{ground}$ when the scanpath has terminated or the referral audio has ended, and $\mL = \mL_{gaze}$ otherwise. Note that $\mathcal{L}_{ground}$ is the auxiliary multi-task grounding loss defined in Sec. 4.2 of the main paper.

To evaluate fixation duration prediction of baselines and ART, we train them on fixation duration prediction along with fixation location prediction. Along with the duration-agnostic metrics we used in the main paper, we also report $SS^{(t)}$, $FED^{(t)}$, $SS_{pack}^{(t)}$, and $FED_{pack}^{(t)}$, which are duration-aware variants (as done in previous works~\cite{cristino2010scanmatch, mondal2023gazeformer, chen2021predicting}) of $SS$, $FED$, $SS_{pack}$, and $FED_{pack}$, respectively. We also report the duration component of MultiMatch~\cite{anderson2015comparison, dewhurst2012depends} ($MM_{dur}$). Higher $SS$, $SS_{pack}$, $SS^{(t)}$, $SS^{(t)}_{pack}$, $CC_{pack}$, $NSS_{pack}$, $MM_{dur}$ metrics signify higher scanpath similarity, whereas higher $FED$, $FED_{pack}$, $FED_{pack}^{(t)}$, and $FED^{(t)}$ metrics denote lower scanpath similarity. The results are in Table~\ref{table:duration_results}. 

ART outperforms baselines on all metrics (both duration-agnostic and duration-aware) when trained on and evaluated for fixation duration prediction and fixation location prediction. The model hyperparameters, pre-training and training processes remain as mentioned in Sec.~\ref{sec:art_details}. Additional details for baselines endowed with fixation prediction can be found in Sec.~\ref{sec:baselines_supp}.

Note that $MM_{dur}$ reflects solely the duration component, in contrast to the spatio-temporal metrics ($SS^{(t)}$, $FED^{(t)}$, $SS_{pack}^{(t)}$, $FED_{pack}^{(t)}$) in Table~\ref{table:duration_results}(b), and by that metric, the random baseline (whose generated fixation duration is set to the average training set fixation duration, as detailed in Sec.~\ref{sec:baselines_supp}) scores higher than the human consistency score. 
We believe this is because of very poor agreement among the behavioral participants in their fixation duration in our incremental object referral task, which makes the prediction of fixation duration less meaningful than the prediction of fixation spatial locations (when we created scanpaths using average fixation locations from the RefCOCO-Gaze training set, we observed $SS = 0.037$ and $SS_{pack}=0.044$, which are far lower than human consistency scores ($SS = 0.400$, $SS_{pack} = 0.317$), signifying that the spatial attention of humans for our task is meaningful and intention-driven).

\def\doubleunderline#1{\underline{\underline{#1}}}
\setlength{\tabcolsep}{3pt}
\begin{table}[bt]
\centering
\caption{Performance of ART and baselines on COCO-Search18~\cite{chen2021coco} dataset. Gazeformer and ART models are shown in two variants - one with fixation prediction capability (``w/ dur.'' in parenthesis) and one without fixation prediction capability (``w/o dur.'' in parenthesis). Metrics in bold are the best performing metrics, while \underline{those} underlined with a single dash are second-best, and  \doubleunderline{those} underlined with a double dash are third-best (we do not underline the third-best metric with double dash for duration-aware metrics since there are only three models predicting fixation duration). }
\label{table:cocosearch18_results}
(a) Duration-agnostic metrics\\
\begin{tabular}{lcccc}
\toprule 
& $SS\bm{\uparrow}$ & $SemSS\bm{\uparrow}$& $FED\bm{\downarrow}$ & $SemFED\bm{\downarrow}$\\  \midrule 
Human & 0.490 & 0.522 & 2.531 & 1.720\\ \midrule
IRL~\cite{yang2020predicting} & 0.405 & 0.441 & 2.781 & 2.393\\
Chen \etal~\cite{chen2021predicting} & 0.398 & 0.425 & 2.376 & \doubleunderline{2.064}\\
FFM~\cite{yang2022target} & 0.384 & 0.391 & 2.719 & 2.479\\
Gazeformer (w/o dur.)~\cite{mondal2023gazeformer} & \textbf{0.475} & \underline{0.456} & \textbf{2.159} & \underline{2.012}\\
Gazeformer(w/ dur.)~\cite{mondal2023gazeformer} & \underline{0.467} & \doubleunderline{0.449} & \underline{2.198} & 2.082\\
\midrule
ART (w/o dur.) (Proposed)& \doubleunderline{0.454} & \textbf{0.461} & \doubleunderline{2.251} & \textbf{1.995}\\
ART(w/ dur.) (Proposed) & 0.432 & 0.441 & 2.335 & 2.070\\
\bottomrule
\end{tabular}\\
\vskip 0.1in
(b) Duration-aware metrics\\
\begin{tabular}{lccccc}
\toprule 
& $SS^{(t)}\bm{\uparrow}$ & $SemSS^{(t)}\bm{\uparrow}$& $FED^{(t)}\bm{\downarrow}$ & $SemFED^{(t)}\bm{\downarrow}$ & $MM_{dur}\bm{\uparrow}$\\  \midrule 
Human & 0.409 & 0.433 & 11.526 & 8.389 & 0.663\\ \midrule
Chen \etal~\cite{chen2021predicting} & 0.354 & 0.368 & 11.610 & 9.991 & 0.691\\
Gazeformer(w/ dur.)~\cite{mondal2023gazeformer} & \textbf{0.417} & \textbf{0.408} & \textbf{10.216} & \textbf{8.771} & \textbf{0.727}\\
\midrule
ART (w/ dur.)& \underline{0.373} & \underline{0.394} & \underline{11.127} & \underline{9.089} & \underline{0.725}\\
\bottomrule
\end{tabular}\\
\end{table}

\section{ART generalizes to Categorical Search (COCO-Search18)}
\label{sec:cocosearch}
In this section, we extend ART to the related categorical search task. We do this via providing a prefix in the form of the category name (\eg, ``car'' or ``potted plant'') to ART. We chose the large-scale categorical search fixation prediction dataset, COCO-Search18~\cite{chen2021coco} to train and evaluate ART and other baselines on its target-present trials. We use several competitive baselines, such as IRL~\cite{yang2020predicting} and FFM~\cite{yang2022target} which are not trained on an additional fixation duration prediction objective, along with Chen~\etal~\cite{chen2021predicting}'s model and Gazeformer~\cite{mondal2023gazeformer} which can be trained on the additional fixation duration prediction objective. State-of-the-art baseline Gazeformer~\cite{mondal2023gazeformer} and ART are trained and evaluated in two variants - one which is trained on the additional fixation duration prediction objective (``w/ dur.'' in parenthesis) and another one which is \textit{not} trained on the additional fixation duration prediction objective (``w/o dur.'' in parenthesis). 

To evaluate on this categorical search task embodied by COCO-Search18, we follow previous methods~\cite{mondal2023gazeformer, yang2022target} and map all predictions to our input grid, and then report Sequence Score~\cite{yang2020predicting} ($SS$), Semantic Sequence Score~\cite{yang2022target} ($SemSS$), Fixation Edit Distance~\cite{mondal2023gazeformer} ($FED$), Semantic Fixation Edit Distance~\cite{mondal2023gazeformer} ($SemFED$) and duration component of MultiMatch~\cite{anderson2015comparison, dewhurst2012depends} ($MM_{dur}$). $SemSS$ and $SemFED$ differs from $SS$ and $FED$, respectively, in that they convert scanpaths to strings of fixated scene object IDs instead of cluster IDs. We also report $SS$, $SemSS$, $FED$, and $SemFED$ with duration denoted by $SS^{(t)}$, $SemSS^{(t)}$, $FED^{(t)}$, and $SemFED^{(t)}$, respectively, as done by previous works~\cite{cristino2010scanmatch, mondal2023gazeformer, chen2021predicting}. Results are in Table~\ref{table:cocosearch18_results}. Higher $SS$, $SemSS$, $SS^{(t)}$, $SemSS^{(t)}$, $MM_{dur}$ metrics signify higher scanpath similarity, whereas higher $FED$, $SemFED$, $FED^{(t)}$, and $SemFED^{(t)}$ metrics denote lower scanpath similarity.

Even though ART is designed for the incremental object referral task, in an extension to the categorical search task, we found that its performance is on par with Gazeformer~\cite{mondal2023gazeformer}, the state-of-the-art search fixation prediction model. ART even outperforms Gazeformer on Semantic Sequence Score (SemSS) and Semantic Fixation Edit Distance (SemFED) metrics. This generalization to the categorical search task further demonstrates the 
strength of ART's architecture.

\section{Additional Details of Baseline Models}
\label{sec:baselines_supp}

In this section, we provide additional implementation details of the baselines used in the main paper. 

\myheading{Random Scanpath}: We sample pack length $l_p$ uniformly from integers [0,1,...,$L_{\mathcal{P}}$] where $L_{\mathcal{P}}$ is the hyperparameter for maximum number of fixations in a pack. Since we chose $L_{\mathcal{P}}=6$ for ART, we use the same value for this baseline for fair comparison. Then we uniformly sample $l_p$ fixation locations within the entire image to obtain a generated pack of fixations. For the variant with fixation duration (Sec.~\ref{sec:durations}), we use the average of all fixation durations in the RefCOCO-Gaze training set.

\myheading{OFA}: We sample pack length $l_p$ uniformly from integers [0,1,...,$L_{\mathcal{P}}$] where $L_{\mathcal{P}}$ is the hyperparameter for maximum number of fixations in a pack. Since we chose $L_{\mathcal{P}}=6$ for ART, we use the same value for this baseline for fair comparison. In order to obtain a generated pack of fixations, we uniformly sample $l_p$ fixation locations from within the bounding box predicted by the OFA~\cite{wang2022ofa} model for the referring expression prefix corresponding to an incoming word within the referring expression. For the variant with fixation duration (Sec.~\ref{sec:durations}), we use the average of all fixation durations in the RefCOCO-Gaze training set.

\myheading{Chen~\etal} We train model from Chen~\etal~\cite{chen2021predicting} using teacher-forcing algorithm~\cite{williams1989learning} in the same manner we have trained ART. To incorporate fixation history, we construct a composite action map containing all previous fixations. We subsequently initialize the dynamic memory of the model with the sum of the task guidance map (from MDETR model which is pre-trained on RefCOCO for fair comparison) and the composite action map. Maximum number of fixations in a predicted pack is set to 6, identical to the value of pack length $L_{\mathcal{P}}$ chosen for ART, for fair comparison. In the context of experiments on RefCOCO-Gaze, we train Chen~\etal's model with fixation duration information for results in Sec.~\ref{sec:durations} and without fixation duration information for the rest, unless specified otherwise.

\myheading{Gazeformer-ref.} We train the Gazeformer~\cite{mondal2023gazeformer} variant, that we named Gazeformer-ref, using teacher-forcing algorithm~\cite{williams1989learning} in the same manner we have trained ART. We provide previous fixation information in the form of the last fixation from the previous non-null pack, which is encoded using a 2D positional encoding and added to the first fixation query as prescribed in \cite{mondal2023gazeformer} for including initial fixation information. The validity prediction head in the model is also extended to support the prediction of an additional end-of-scanpath token. Maximum number of fixations in a predicted pack is set to 6, identical to the value of pack length $L_{\mathcal{P}}$ chosen for ART, for fair comparison. In the context of experiments on RefCOCO-Gaze, we train Gazeformer-ref with fixation duration information for results in Sec.~\ref{sec:durations} and without fixation duration information for the rest, unless specified otherwise.

\myheading{Gazeformer-cat.} We train the Gazeformer~\cite{mondal2023gazeformer} variant, that we named Gazeformer-cat, using teacher-forcing algorithm~\cite{williams1989learning} in the same manner we have trained ART. The previous fixation history is conveyed in the same manner as in the implementation of Gazeformer-ref (see above). We also extend the validity prediction head to support the prediction of an additional end-of-scanpath token. The target category estimation which is used to construct the input category name comes from a RoBERTa-based classifier which is separately trained on RefCOCO referring expressions and their corresponding target categories. Specifically, the target category estimator is a RoBERTa-base model with a classification head on top. This baseline should show how important target category estimation is for gaze prediction. Maximum number of fixations in a predicted pack is set to 6, identical to the value of pack length $L_{\mathcal{P}}$ chosen for ART, for fair comparison. In the context of experiments on RefCOCO-Gaze, we train Gazeformer-cat with fixation duration information for results in Sec.~\ref{sec:durations} and without fixation duration information for the rest, unless specified otherwise.

\section{Additional metrics for Ablation Studies}
\label{sec:ablations_comp}
\setlength{\tabcolsep}{3pt}
\begin{table*}[bt]
\centering
\caption{{\bf Ablation studies on ART model (reported in Table~\ref{table:ablate_results} of the main paper) augmented with additional metrics.} If either $\mathcal{L}_{bbox}$ or $\mathcal{L}_{target}$ is included, and the model undergoes pre-training, the loss is applied in \textit{both} pre-training and gaze training phases.}
\begin{tabular}{c|ccc|cccccc}
\toprule 
Ablation & Pre- & $\mathcal{L}_{bbox}$ & $\mathcal{L}_{target}$ & $SS$ & $SS_{pack}$& $FED$ & $FED_{pack}$&$CC_{pack}$& $NSS_{pack}$\\
\# & training &  &  & $\bm{\uparrow}$ & $\bm{\uparrow}$ & $\bm{\downarrow}$ & $\bm{\downarrow}$ & $\bm{\uparrow}$ & $\bm{\uparrow}$ \\  
\midrule 

1 &  $\times$ & $\times$ & $\times$ & 0.309 &  0.257 & 6.873 & 1.203 &  0.222 &  3.032\\
2 &  $\checkmark$ & $\checkmark$ & $\times$ &  0.321 &  0.279 &  7.341 & 1.348 & 0.239 &  2.769\\
3 & $\checkmark$ & $\times$ & $\checkmark$ & 0.292  & 0.260 & 6.713 & 1.162 & 0.216 &  2.967\\
4 & $\times$& $\checkmark$  & $\checkmark$ & 0.304  & 0.257 & 7.104 & 1.245& 0.215 & 2.953 \\
5 & $\checkmark$& $\checkmark$  & $\checkmark$ & \textbf{0.359}  & \textbf{0.292} &  \textbf{6.371} & \textbf{1.143} & \textbf{0.280} & \textbf{3.478} \\
\bottomrule
\end{tabular}

\vspace{-0.1in}
\label{table:ablate_results_complete}
\end{table*}

In Table~\ref{table:ablate_results_complete}, we augment Table~\ref{table:ablate_results} in the main paper with additional metrics ($FED$, $FED_{pack}$ and $NSS_{pack}$). The trends remain similar to the what we observed for $SS$ and $SS_{pack}$ scores, thereby reaffirming our assertion that \textit{both} object localization and target category prediction tasks are integral to the object referral process and that pre-training on these tasks is instrumental for superior performance.

\section{Additional Ablation Studies and Analysis}
\label{sec:ablations_supp}

\setlength{\tabcolsep}{3pt}
\begin{table}[bt]
\centering
\caption{{\bf Additional ablation studies on ART model.} If either $\mathcal{L}_{bbox}$ or $\mathcal{L}_{target}$ is included, and the model undergoes pre-training, the loss is applied in \textit{both} pre-training and gaze training phases. Ablations \#3, \#4 and \#5 are from Table~\ref{table:ablate_results_complete} (also in Table~\ref{table:ablate_results} in the main paper). }
\begin{tabular}{c|ccc|cccccc}
\toprule 
Ablation & Pre- & $\mathcal{L}_{bbox}$ & $\mathcal{L}_{target}$ & $SS$ & $SS_{pack}$& $FED$ & $FED_{pack}$&$CC_{pack}$& $NSS_{pack}$\\
\# & training &  &  & $\bm{\uparrow}$ & $\bm{\uparrow}$ & $\bm{\downarrow}$ & $\bm{\downarrow}$ & $\bm{\uparrow}$ & $\bm{\uparrow}$ \\ 
\midrule 
1 &  $\times$ & $\checkmark$ & $\times$ & 0.319 & 0.270 & 6.736 & 1.193 & 0.228 & 3.135\\
2 &  $\times$ & $\times$ & $\checkmark$ &   0.309& 0.264 & 6.705 & 1.175 &  0.216 & 2.919\\
3 &  $\checkmark$ & $\checkmark$ & $\times$ &  0.321 &  0.279 &  7.341 & 1.348 &  0.239 &  2.769\\
4 & $\checkmark$ & $\times$ & $\checkmark$ & 0.292  & 0.260 & 6.713 & 1.162 & 0.216 &  2.967\\
5 & $\checkmark$& $\checkmark$  & $\checkmark$ & \textbf{0.359}  & \textbf{0.292} &   \textbf{6.371} & \textbf{1.143} & \textbf{0.280} & \textbf{3.478} \\
\bottomrule
\end{tabular}
\label{table:ablate_results_suppl}

\end{table}

We provide two additional ablations (in addition to the five ablations in Table~\ref{table:ablate_results} of the main paper and Table~\ref{table:ablate_results_complete} in Section~\ref{sec:ablations_comp}) tabulated as ablation \#1 and ablation \#2 in Table~\ref{table:ablate_results_suppl}. As it can be seen, addition of only one of the auxiliary losses $\mathcal{L}_{bbox}$ and $\mathcal{L}_{target}$ results in little to no boost in performance. It is evident that we need \textit{both} $\mathcal{L}_{bbox}$ and $\mathcal{L}_{target}$ in the objective function \textit{along with pre-training} for our model to achieve high performance. We posit that ablation \#4 in Table~\ref{table:ablate_results_suppl} fails to perform well because the target category prediction task largely, if not completely, relies on the linguistic input (i.e., the referring expression). Consequently, the sub-networks dedicated to visual and visuo-linguistic processing (especially, the visual encoder) might not benefit from pre-training only on this objective whereas the linguistic subnetworks (i.e. the linguistic encoder) are greatly optimized. 
We speculate that it is also perhaps hard for ART to adapt to object referral during training after pre-training its parameters to significantly align with the target-category estimation objective (which can be inadequate for our task since there are multiple objects belonging to the target category) in ablation \#4 of Table~\ref{table:ablate_results_suppl}. On the other hand, object localization seems to be much more aligned with the object referral task, which is indeed shown by ablation \#3 in Table~\ref{table:ablate_results_suppl}. In summary, the ablation studies validate our hypothesis that \textit{both} object localization and target category prediction tasks are integral to the object referral process. Also note that Ablations 1 and 4 in Table~\ref{table:ablate_results_complete} and Ablations 1 and 2 in Table~\ref{table:ablate_results_suppl} show that even when ART is not pre-trained on RefCOCO, it still outperforms baselines that are also not pre-trained on RefCOCO, i.e., Gazeformer-ref and Gazeformer-cat (in Table~\ref{table:main_results} of main text).

\setlength{\tabcolsep}{3pt}
\begin{table}[bt]
\centering
\caption{{\bf Additional ablation studies on ART model when trained with additional Fixation Duration Prediction objective.} If either $\mathcal{L}_{bbox}$ or $\mathcal{L}_{target}$ is included, and the model undergoes pre-training, the loss is applied in \textit{both} pre-training and gaze training phases. }

(a) Duration-agnostic Metrics
\begin{tabular}{c|ccc|cccccc}
\toprule 
Ablation & Pre- & $\mathcal{L}_{bbox}$ & $\mathcal{L}_{target}$ & $SS$ & $SS_{pack}$& $FED$ & $FED_{pack}$&$CC_{pack}$& $NSS_{pack}$\\
\# & training &  &  & $\bm{\uparrow}$ & $\bm{\uparrow}$ & $\bm{\downarrow}$ & $\bm{\downarrow}$ & $\bm{\uparrow}$ & $\bm{\uparrow}$ \\ 
\midrule 
1 &  $\times$ & $\times$ & $\times$ & 0.206 & 0.169 & 10.840 & 2.073 & 0.167 & 2.331\\
2 &  $\times$ & $\checkmark$ & $\times$ & 0.262 & 0.230 & 8.225 & 1.528 & 0.207 & 2.930\\
3 &  $\times$ & $\times$ & $\checkmark$ & 0.269 & 0.201 & 9.626 & 1.616 & 0.196 & 2.396\\
4 &  $\times$ & $\checkmark$ & $\checkmark$ & 0.296 & 0.252 & 7.049 &  1.271& 0.209 &2.914\\
5 &  $\checkmark$ & $\checkmark$ & $\times$ & 0.309  & 0.278 & 7.306 & 1.339 & 0.257 & 3.307\\
6 & $\checkmark$ & $\times$ & $\checkmark$ & 0.284 & 0.210 & 7.172 &  1.351 & 0.179 & 2.460\\
7 & $\checkmark$& $\checkmark$  & $\checkmark$ &  \textbf{0.356} & \textbf{0.285} &  \textbf{6.410} & \textbf{1.161} & \textbf{0.281} & \textbf{3.539} \\
\bottomrule
\end{tabular}
\vskip 0.2in
(b) Duration-aware Metrics
\begin{tabular}{c|ccc|ccccc}
\toprule 
Ablation & Pre- & $\mathcal{L}_{bbox}$ & $\mathcal{L}_{target}$ & $SS^{(t)}$ & $SS_{pack}^{(t)}$& $FED^{(t)}$ & $FED_{pack}^{(t)}$& $MM_{dur}$\\
\# & training &  &  & $\bm{\uparrow}$ & $\bm{\uparrow}$ & $\bm{\downarrow}$ & $\bm{\downarrow}$ & $\bm{\uparrow}$  \\ 
\midrule 
1 &  $\times$ & $\times$ & $\times$ & 0.222 & 0.110 & 55.614 &  11.274 & 0.672\\
2 &  $\times$ & $\checkmark$ & $\times$ & 0.253 & 0.169 & 44.074 & 8.559  & 0.691\\
3 &  $\times$ & $\times$ & $\checkmark$ & 0.277 & 0.159 & 47.876 & 8.928  & 0.675\\
4 &  $\times$ & $\checkmark$ & $\checkmark$ & 0.253 & 0.164 & 38.722 & 7.232 & 0.652\\
5 &  $\checkmark$ & $\checkmark$ & $\times$ & 0.282 & 0.181 & 37.922 & 7.389  & 0.685\\
6 & $\checkmark$ & $\times$ & $\checkmark$ & 0.252 & 0.165 & 38.934 &7.206  & 0.683\\
7 & $\checkmark$& $\checkmark$  & $\checkmark$ & \textbf{0.332} & \textbf{0.199} & \textbf{35.997} & \textbf{7.120} & \textbf{0.696} \\
\bottomrule
\end{tabular}

\label{table:ablate_results_duration}
\end{table}

In Table~\ref{table:ablate_results_duration}, we tabulate the ablation studies for ART when equipped with fixation duration prediction capability. As shown in Sec.~\ref{sec:durations} through analysis of $MM_{dur}$ metric values for random baseline and human consistency, there is poor agreement between participants in their fixation durations and thus training on such noisy supervision can be challenging. We interpret Table~\ref{table:ablate_results_duration} as being consistent with our findings from the ablation studies with ART without fixation prediction (Sec. 5.4 in main text;  Sec.~\ref{sec:ablations_comp}, Sec.~\ref{sec:ablations_supp} in the supplement) in supporting our assertion that \textit{both} pre-training and training on \textit{both} auxiliary object localization and target category estimation objectives are crucial for ART's performance. We also observe that without pre-training and training on auxiliary losses, ART struggles to generalize when trained with noisy fixation durations, as seen in Ablation 1. Our proposed model (\#7 in Table~\ref{table:ablate_results_duration}) supports our assertion that ART's pre-training and training on the two object grounding tasks for complete/partial expressions underlying object referral significantly contributes towards its SOTA performance when compared to existing baselines that are unable to pre-train/train on object grounding for partial/complete expressions. Our proposed model thus generalizes well even when trained with a noisy supervision signal, such as the fixation durations. We also note that when not pre-trained on RefCOCO (Ablations 2, 3, and 4 in Table~\ref{table:ablate_results_duration}), ART still outperforms baselines that are not pre-trained on RefCOCO, i.e. Gazeformer-ref and Gazeformer-cat (see Table~\ref{table:duration_results}).

\section{Auxiliary Next Word Token Prediction Task}
\label{sec:next_word_supp}

We also hypothesized that predicting the next linguistic token during search also underlies incremental object referral process along with the object localization and target category prediction tasks. So we added a \texttt{NEXT\_WORD\_TOKEN} token along with \texttt{BBOX} and \texttt{TGT} tokens as input to the visuo-linguistic encoder. The corresponding latent vector served as input to an MLP which generated logits over a vocabulary of tokens in order to predict the next word token. The loss imposed is a cross-entropy loss $\mathcal{L}_{nextword}$. The results are tabulated in Table~\ref{table:nextword_suppl}. As we can see, adding $\mathcal{L}_{nextword}$ does not improve the performance significantly -- we achieve best performance without $\mathcal{L}_{nextword}$ (Ablation \#1 in Table~\ref{table:nextword_suppl}). We hypothesize that this is because the next word token prediction task is considerably more difficult than the object localization and target category prediction tasks, and potentially introduces noise while training on the gaze prediction objective.

\setlength{\tabcolsep}{3pt}
\begin{table}[ht!]
\centering
\caption{{\bf Effect of auxiliary next word token prediction task on ART model.} Ablations \#1 and \#3 are from Table~\ref{table:ablate_results} in main text. }
\begin{tabular}{c|cccc|cccc}
\toprule 
Sl.No\#&Pre-training & $\mathcal{L}_{nextword}$ & $\mathcal{L}_{bbox}$  & $\mathcal{L}_{target}$ & $SS$ & $SS_{pack}$ & $CC_{pack}$& $NSS_{pack}$\\  \midrule 
1 &\checkmark & $\times$ & $\checkmark$  & $\checkmark$ & \textbf{0.359}  & \textbf{0.292} & \textbf{0.280} & \textbf{3.478} \\
2 & \checkmark &  $\checkmark$ & $\checkmark$ & $\checkmark$ & 0.355 & 0.281 & 0.269 & 3.388\\
3 & $\times$ &  $\times$ & $\checkmark$ & $\checkmark$ & 0.304 & 0.257 & 0.215 & 2.953\\
4 & $\times$ &  $\checkmark$ & $\checkmark$ & $\checkmark$ & 0.313 & 0.265 & 0.224 & 3.049\\

\bottomrule
\end{tabular}
\label{table:nextword_suppl}
\end{table}


\section{Qualitative Results}
\label{sec:qual_supp}

In this section, we present additional qualitative results of human behavior, our model ART and other competitive baseline models in Fig.~\ref{fig:qualitative_results} and Fig.~\ref{fig:qualitative_results_2}. We see that ART efficiently finds the correct target through scanpaths that closely resemble human behavior in all rows except the last row in Fig.~\ref{fig:qualitative_results_2} - where it fails to localize the ``head'' of the correct person. In the first row of Fig.~\ref{fig:qualitative_results}, we can see both the human participant and ART \textit{wait} until the last word ``left'' to localize the correct muffin. We see a similar \textit{waiting} pattern in second and third row of Fig.~\ref{fig:qualitative_results} where ART and the human participant wait until disambiguation towards the end of the expression for localizing the correct ``bus'' and ``car'' respectively. In the first row of Fig.~\ref{fig:qualitative_results_2}, we see a \textit{scanning} behavior where ART and the human fixate on the kids in the center after hearing the word ``kid'' until the contextual information ``far right'' is provided in the end to locate the correct ``kid''. In the second row of Fig.~\ref{fig:qualitative_results_2}, we observe that both ART and the human participant \textit{wait} till the utterance of the target category word ``elephant'' in order to localize the correct elephant.

\def\subFigSz{0.23\linewidth}
\begin{figure}[ht!]
\centering

\includegraphics{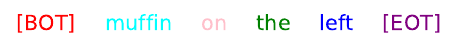} \\
  \includegraphics[width=\subFigSz]{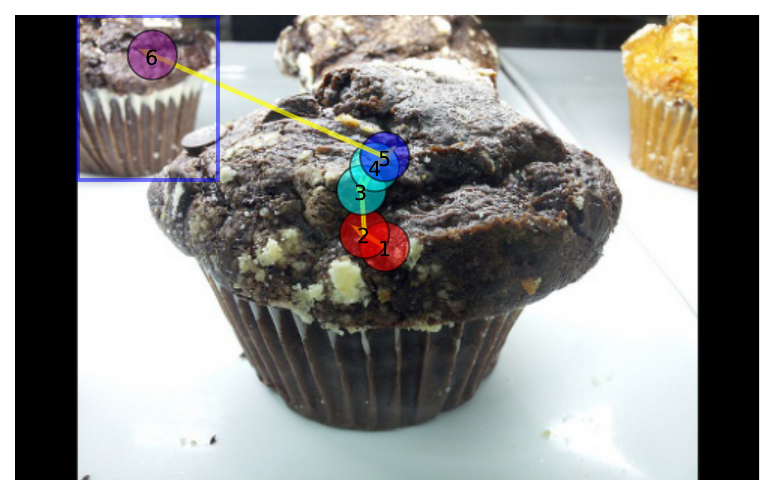} 
 \includegraphics[width=\subFigSz]{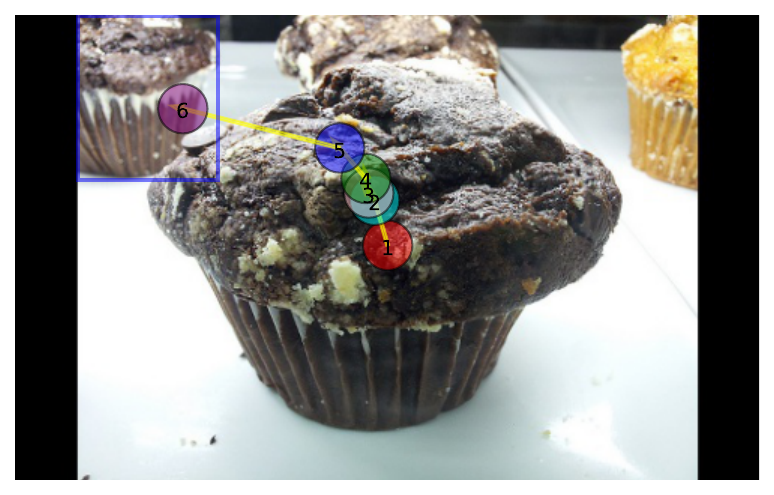}  
  \includegraphics[width=\subFigSz]{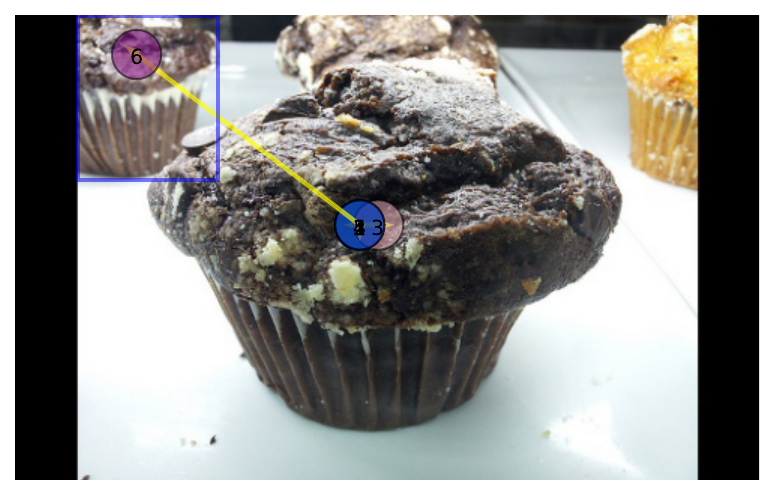} 
  \includegraphics[width=\subFigSz]{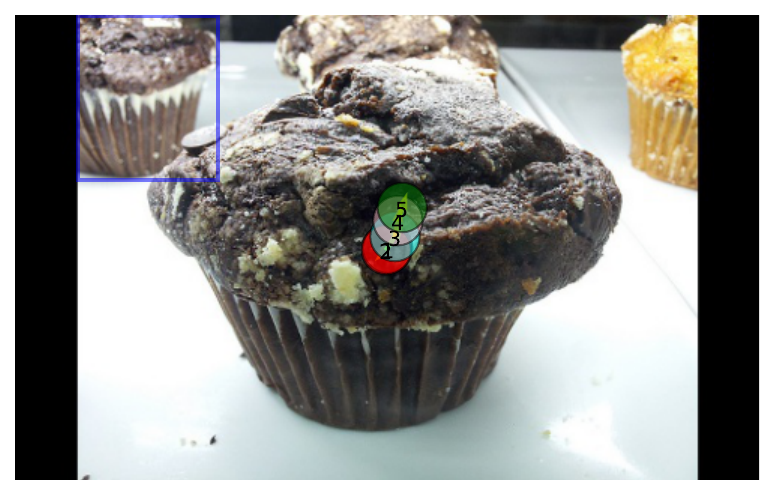} \makebox[\subFigSz]{\small{Human}}
\makebox[\subFigSz]{\small{ART (Proposed)}}
\makebox[\subFigSz]{\small{Chen \etal}}
\makebox[\subFigSz]{\small{Gazeformer-ref}}\\
\includegraphics{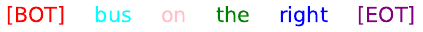} \\
  \includegraphics[width=\subFigSz]{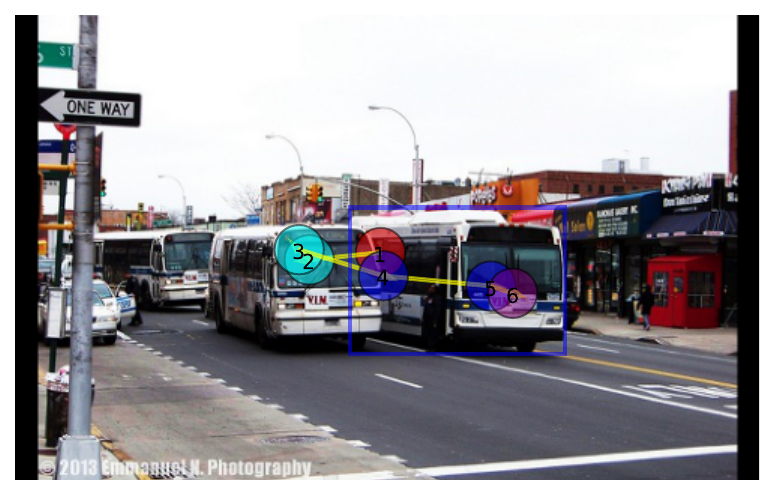} 
 \includegraphics[width=\subFigSz]{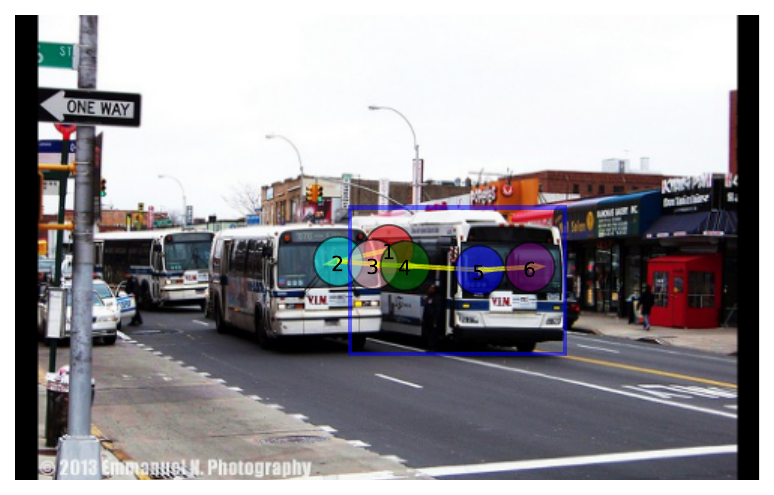}  
  \includegraphics[width=\subFigSz]{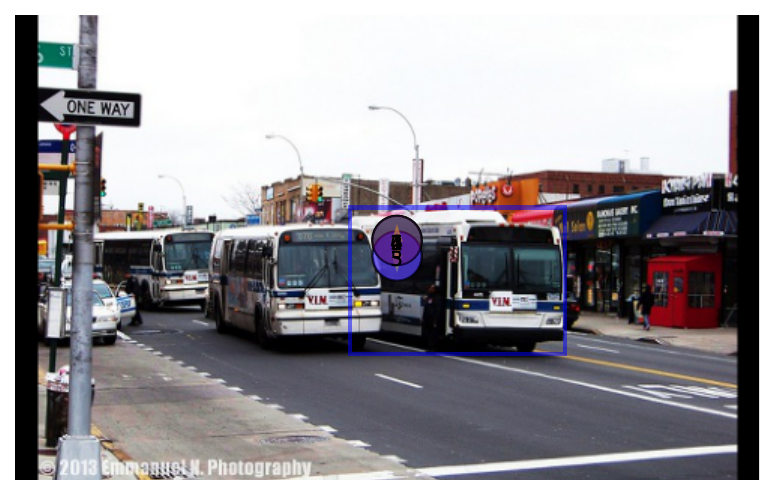} 
  \includegraphics[width=\subFigSz]{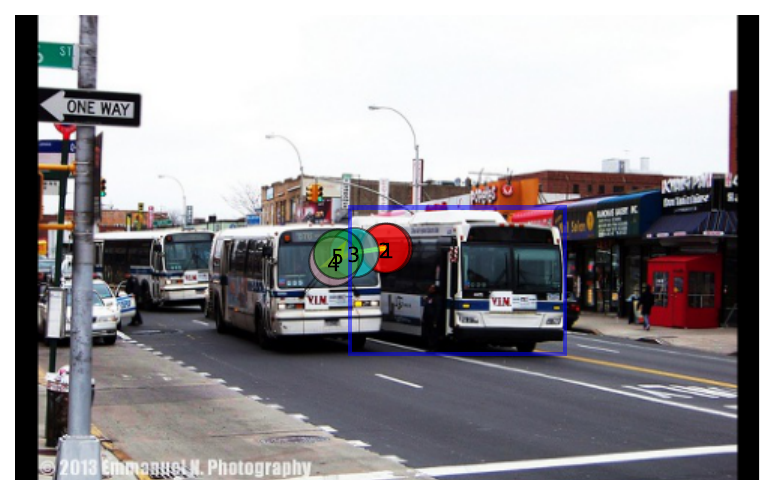} \makebox[\subFigSz]{\small{Human}}
\makebox[\subFigSz]{\small{ART (Proposed)}}
\makebox[\subFigSz]{\small{Chen \etal}}
\makebox[\subFigSz]{\small{Gazeformer-ref}}\\
\includegraphics{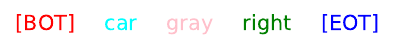} \\
  \includegraphics[width=\subFigSz]{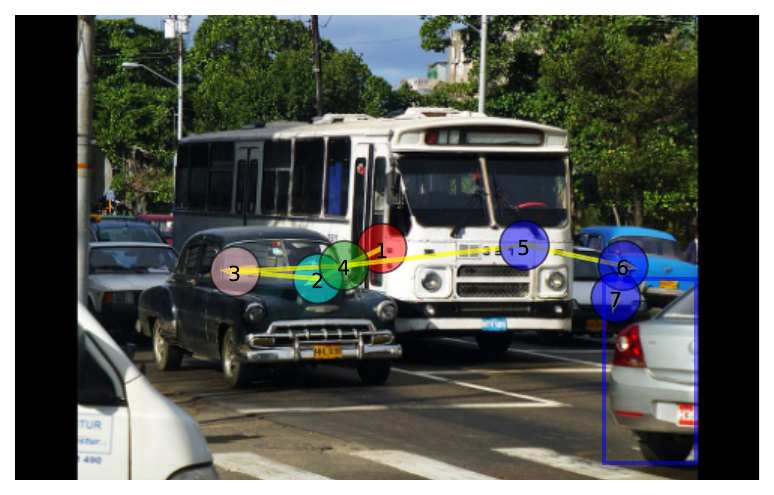} 
 \includegraphics[width=\subFigSz]{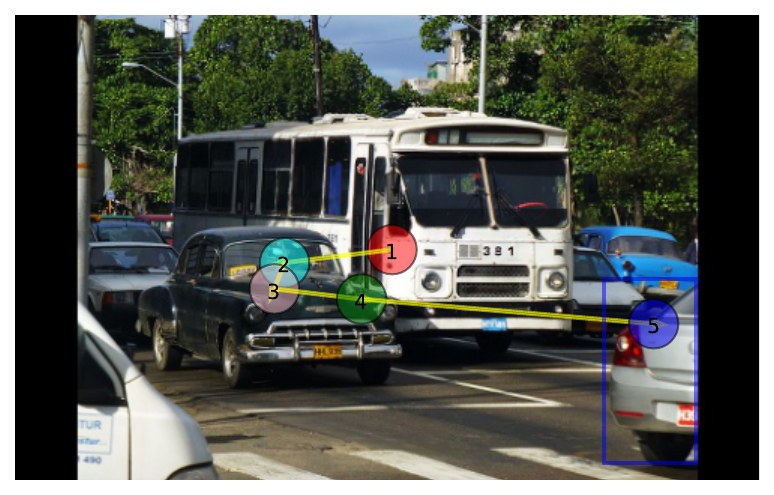}  
  \includegraphics[width=\subFigSz]{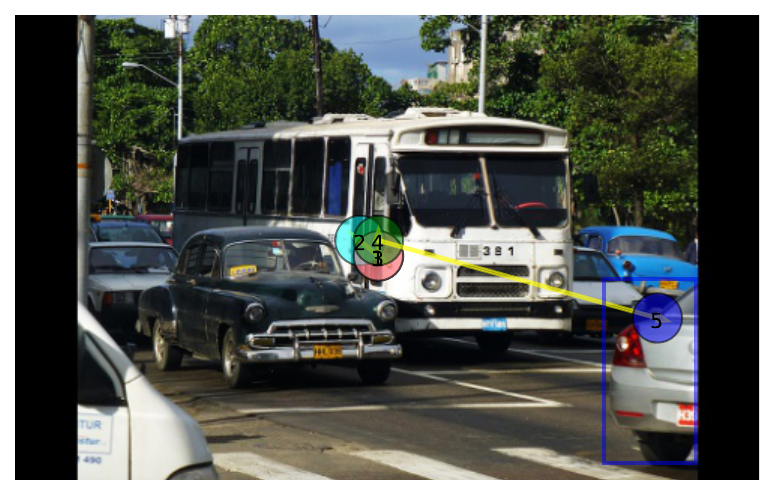} 
  \includegraphics[width=\subFigSz]{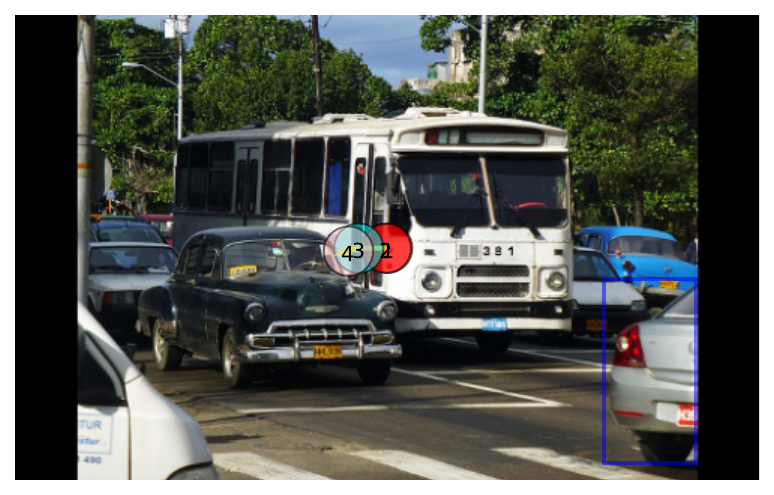} \makebox[\subFigSz]{\small{Human}}
\makebox[\subFigSz]{\small{ART (Proposed)}}
\makebox[\subFigSz]{\small{Chen \etal}}
\makebox[\subFigSz]{\small{Gazeformer-ref}}\\
\caption{{\bf Qualitative results [1/2].}  Scanpaths from humans and three scanpath prediction models on three trials exhibiting strategic fixation behavior. Fixations (denoted by circles numbered with fixation order) are color-coded to corresponding words in the referring expression (above each row). Fixations color-coded to \texttt{[BOT]} occurred before the expression started, and those color-coded to \texttt{[EOT]} occurred after the expression ended. Blue bounding boxes indicating the referred objects are not visible during trials. Our model generates the most human-like scanpaths for incremental object referral.}
\label{fig:qualitative_results}
\end{figure}

\def\subFigSz{0.23\linewidth}
\begin{figure}[ht!]
\centering

\includegraphics{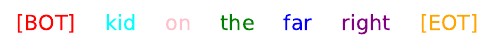} \\
  \includegraphics[width=\subFigSz]{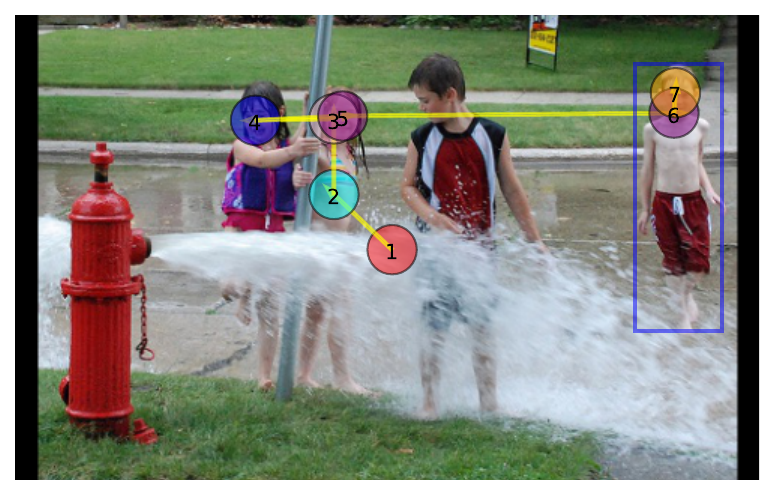} 
 \includegraphics[width=\subFigSz]{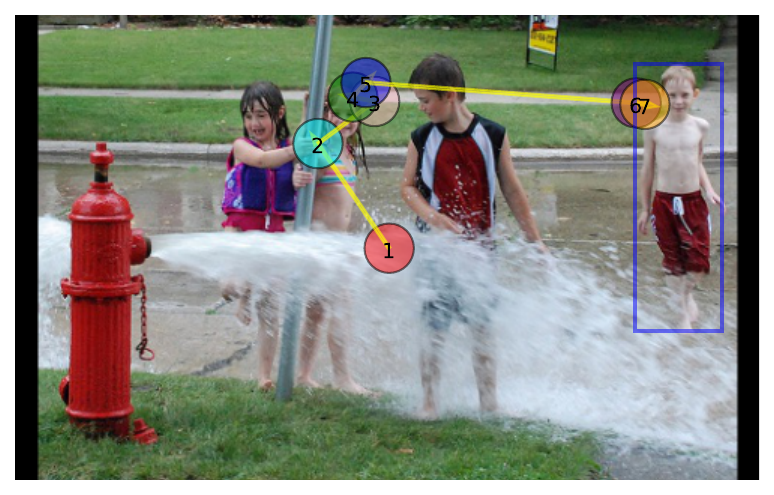}  
  \includegraphics[width=\subFigSz]{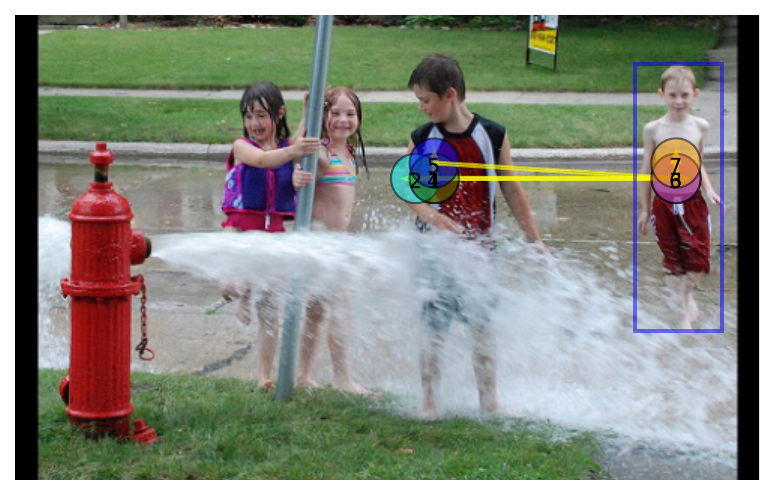} 
  \includegraphics[width=\subFigSz]{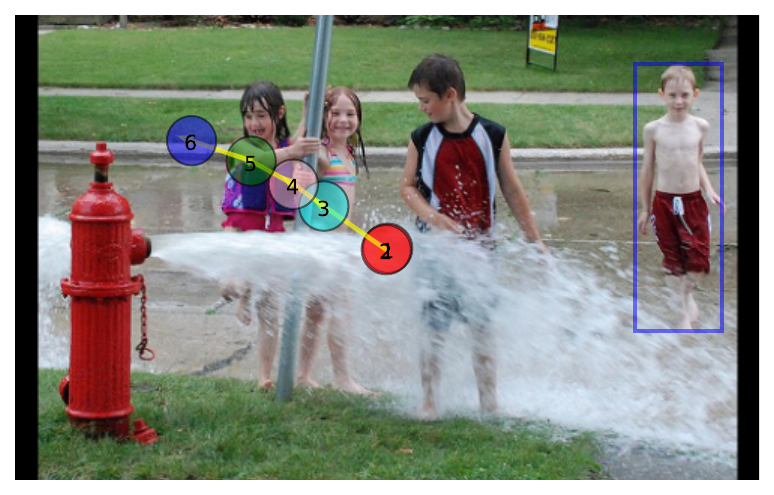} \makebox[\subFigSz]{\small{Human}}
\makebox[\subFigSz]{\small{ART (Proposed)}}
\makebox[\subFigSz]{\small{Chen \etal}}
\makebox[\subFigSz]{\small{Gazeformer-ref}}\\
\includegraphics{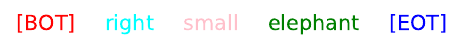} \\
  \includegraphics[width=\subFigSz]{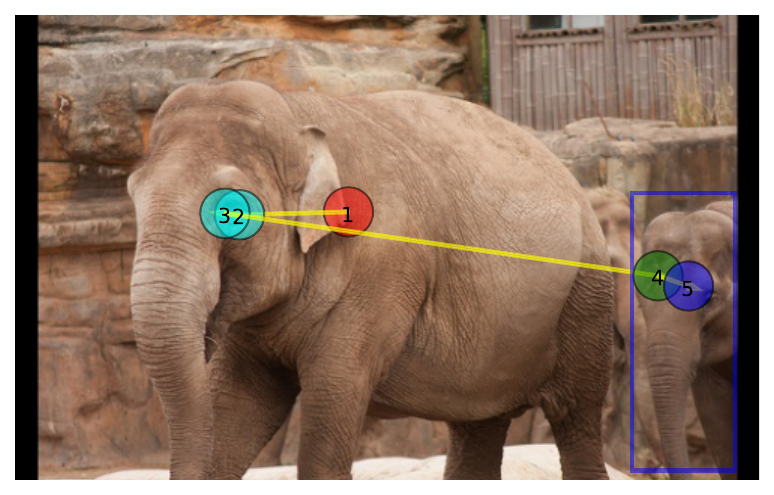} 
 \includegraphics[width=\subFigSz]{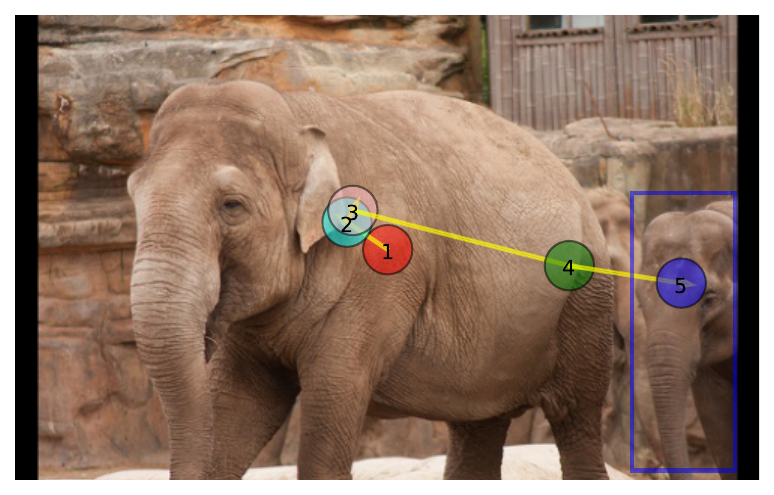}  
  \includegraphics[width=\subFigSz]{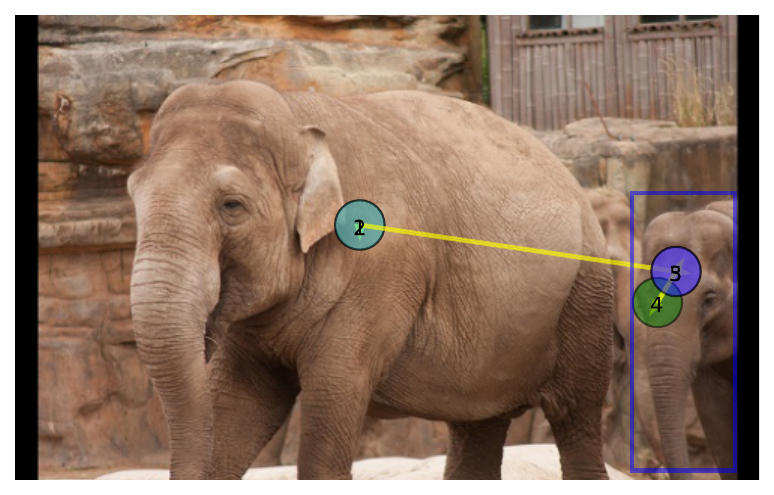} 
  \includegraphics[width=\subFigSz]{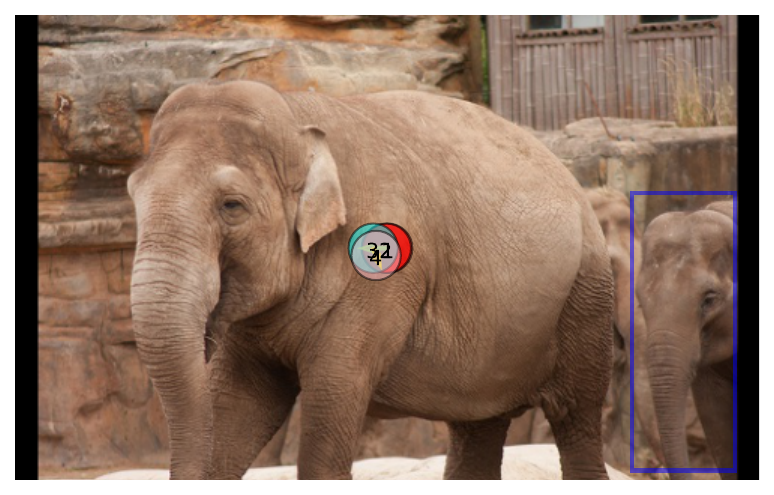} \makebox[\subFigSz]{\small{Human}}
\makebox[\subFigSz]{\small{ART (Proposed)}}
\makebox[\subFigSz]{\small{Chen \etal}}
\makebox[\subFigSz]{\small{Gazeformer-ref}}\\

\includegraphics{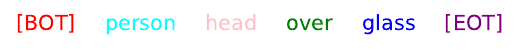} \\
  \includegraphics[width=\subFigSz]{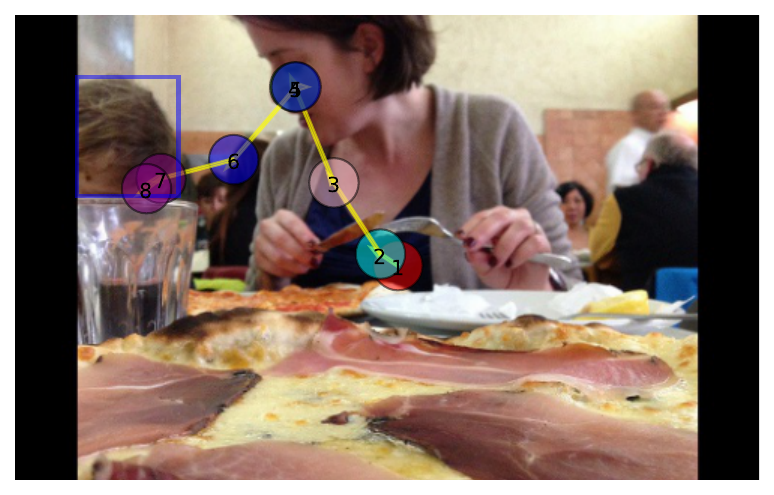} 
 \includegraphics[width=\subFigSz]{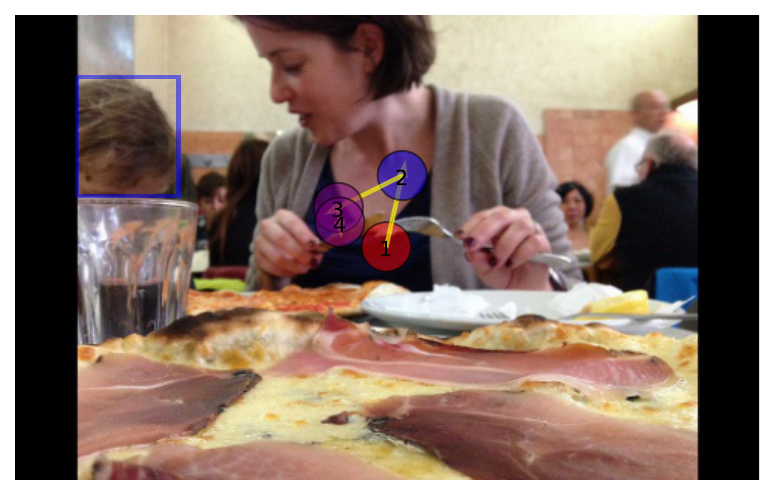}  
  \includegraphics[width=\subFigSz]{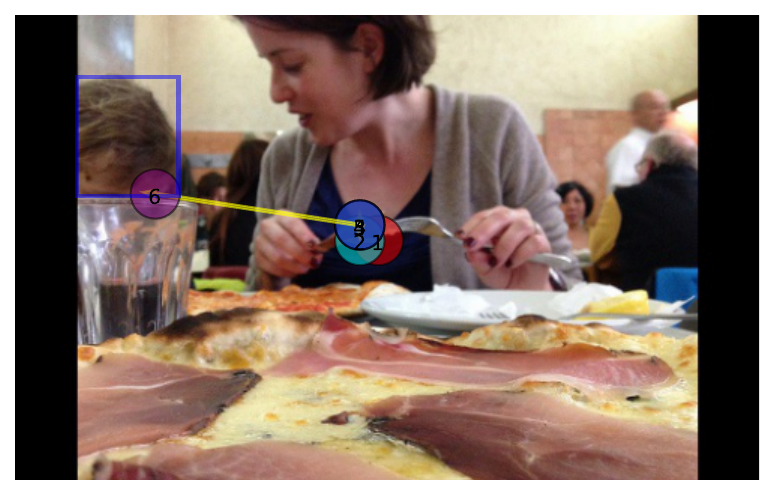} 
  \includegraphics[width=\subFigSz]{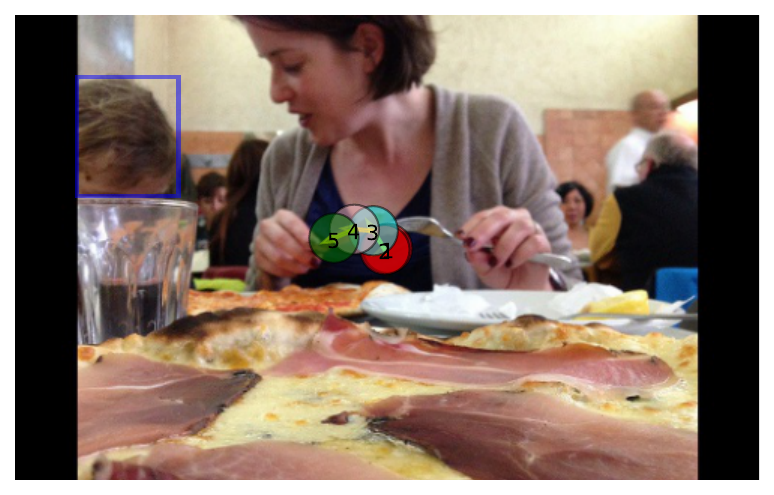} \makebox[\subFigSz]{\small{Human}}
\makebox[\subFigSz]{\small{ART (Proposed)}}
\makebox[\subFigSz]{\small{Chen \etal}}
\makebox[\subFigSz]{\small{Gazeformer-ref}}\\
\caption{{\bf Qualitative results [2/2].}  Scanpaths from humans and three scanpath prediction models on three trials exhibiting strategic fixation behavior. Fixations (denoted by circles numbered with fixation order) are color-coded to corresponding words in the referring expression (above each row). Fixations color-coded to \texttt{[BOT]} occurred before the expression started, and those color-coded to \texttt{[EOT]} occurred after the expression ended. Blue bounding boxes indicating the referred objects are not visible during trials. Our model generates the most human-like scanpaths for incremental object referral.}
\label{fig:qualitative_results_2}
\end{figure}
\clearpage

\bibliographystyle{splncs04}
\bibliography{main}
\end{document}

\bibliographystyle{splncs04}
\bibliography{main}
\end{document}